\documentclass[journal]{IEEEtran}

\usepackage[utf8]{inputenc}
\usepackage{xcolor}
\usepackage{amsmath}
\usepackage{graphicx}
\usepackage{times}
\usepackage{soul}
\usepackage[small]{caption}
\usepackage{amsfonts}
\usepackage{array}
\usepackage{multirow}
\usepackage{longtable}
\usepackage{subfigure}
\usepackage{algpseudocode}
\usepackage[linesnumbered,ruled]{algorithm2e}

\usepackage{cite}
\usepackage{pifont}% http://ctan.org/pkg/pifont
\usepackage{arydshln} % for cdashline
\usepackage[normalem]{ulem}

\newcommand{\cmark}{\ding{51}}%
\newcommand{\xmark}{\ding{55}}%

\def\BB{\textcolor{black}}

\def\eg{\emph{e.g.}} 
\def\ie{\emph{i.e.}}

\usepackage{amsmath,amsfonts,bm}

\def\eqref#1{equation~\ref{#1}}
\def\1{\bm{1}}
\newcommand{\train}{\mathcal{D}}

\def\rvx{{\mathbf{x}}}
\def\rvy{{\mathbf{y}}}

\def\rmC{{\mathbf{C}}}

\def\rmI{{\mathbf{I}}}

\def\rmO{{\mathbf{O}}}

\def\rmR{{\mathbf{R}}}

\DeclareMathAlphabet{\mathsfit}{\encodingdefault}{\sfdefault}{m}{sl}
\SetMathAlphabet{\mathsfit}{bold}{\encodingdefault}{\sfdefault}{bx}{n}

\def\gD{{\mathcal{D}}}

\def\gF{{\mathcal{F}}}

\newcommand{\etens}[1]{\mathsfit{#1}}

\def\etN{{\etens{N}}}

\newcommand{\R}{\mathbb{R}}

\begin{document}
%\title{Dynamic and Asymptotic Filter Pruning for Efficient Neural Networks}
%\title{Asymptotic Soft Filter Pruning for ConvNets}
\title{Asymptotic Soft Filter Pruning\\for Deep Convolutional Neural Networks}
%Progressive Acceleration Efficient inference

\author{Yang~He,
        Xuanyi~Dong,
        Guoliang~Kang, 
        Yanwei~Fu,
        Chenggang~Yan,
        and~Yi~Yang{$^*$}
%\author{Yang~He,~\IEEEmembership{Member,~IEEE,}
%        John~Doe,~\IEEEmembership{Fellow,~OSA,}
%        and~Jane~Doe,~\IEEEmembership{Life~Fellow,~IEEE}% <-this % stops a space
\thanks{
Y. He, X.~Dong, G.~Kang and Y.~Yang are with the Center for AI, University of Technology Sydney, Sydney,
NSW 2007, Australia (e-mail: yang.he-1@student.uts.edu.au; xuanyi.dong@student.uts.edu.au; guoliang.kang@student.uts.edu.au; yi.yang@uts.edu.au.).

Y. Fu is with The School of Data Science, Fudan University, Shanghai 200433, China (e-mail: yanweifu@fudan.edu.cn.) 

C. Yan is with Hangdian University, Hangzhou 310000, China (cgyan@hdu.edu.cn)
}
\thanks{{*}Corresponding Author: Yi Yang (yi.yang@uts.edu.au)}
% <-this % stops a space
%\thanks{Y. He and Y.~Yang are also with the SUSTech-UTS Joint Centre of CIS (SUCCIS), Southern University of Science and Technology, Guangdong 518005, China.}% <-this % stops a space
}

\markboth{Journal of \LaTeX\ Class Files,~Vol.~14, No.~8, August~2015}%
{Shell \MakeLowercase{\textit{et al.}}: Bare Demo of IEEEtran.cls for IEEE Journals}



\maketitle

\begin{abstract}
Deeper and wider Convolutional Neural Networks (CNNs) achieve superior performance but bring expensive computation cost. 
Accelerating such over-parameterized neural network has received increased attention.
A typical pruning algorithm is a three-stage pipeline, i.e., training, pruning, and retraining.
Prevailing approaches fix the pruned filters to zero during retraining, and thus significantly reduce the optimization space. 
Besides, they directly prune a large number of filters at first, which would cause unrecoverable information loss.
To solve these problems, we propose an Asymptotic Soft Filter Pruning (ASFP) method to accelerate the inference procedure of the deep neural networks.
First, we update the pruned filters during the retraining stage. As a result, the optimization space of the pruned model would not be reduced but be the same as that of the original model. In this way, the model has enough capacity to learn from the training data.
Second, we prune the network asymptotically. We prune few filters at first and asymptotically prune more filters during the training procedure.
With asymptotic pruning, the information of the training set would be gradually concentrated in the remaining filters, so the subsequent training and pruning process would be stable.
Experiments show the effectiveness of our ASFP on image classification benchmarks.
Notably, on ILSVRC-2012, our ASFP reduces more than 40\% FLOPs on ResNet-50 with only 0.14\% top-5 accuracy degradation, which is higher than the soft filter pruning (SFP) by 8\%.
\end{abstract}

\begin{IEEEkeywords}
Filter Pruning, Image Classification, Neural Networks
\end{IEEEkeywords}

% \fi
\IEEEpeerreviewmaketitle

\section{Introduction}

\IEEEPARstart{C}{onvolutional} Neural Networks (CNNs) have demonstrated state-of-the-art performance in computer vision tasks~\cite{krizhevsky2012imagenet,yang2015multitask,ren2015faster,chang2017bi,dong2019search,luo2018adaptive,du2017stacked,wei2017cross,zhang2017visual,wu2019progressive,dong2019hcmf,kang2017shakeout}.
The superior performance of deep CNNs usually comes from the deeper and wider architectures~\cite{krizhevsky2012imagenet,simonyan2014very,szegedy2015going,szegedy2016rethinking,he2016deep}, which cause the prohibitively expensive computation cost.
%Even if we use more efficient architectures, such as residual connections~\cite{he2016deep} or inception modules~\cite{szegedy2016rethinking}, it is still difficult in deploying the state-of-the-art CNN models on mobile devices.
The storage, memory, and computation of these cumbersome models significantly exceed the computing limitation of current mobile devices or drones.
\cite{han2015learning} shows that running a 1 billion connection neural network at 20Hz would require 12.8W just for DRAM access.
%, which beyond the power envelope of a typical mobile device.
Besides, the authors of \cite{chen2016eyeriss,sze2017efficient} claim that  VGGNet~\cite{simonyan2014very} requires 321.1 MBytes memory and 236 mW power for a batch of three frames and processes 0.7 frame per second even when it is deployed on the optimized energy-efficient chip.
Therefore, it is essential to maintain the deep CNN models to have a relatively low computational cost but ensure high accuracy in real-world applications.

\begin{figure}[!t]
\begin{centering}
\includegraphics[width=0.9\linewidth]{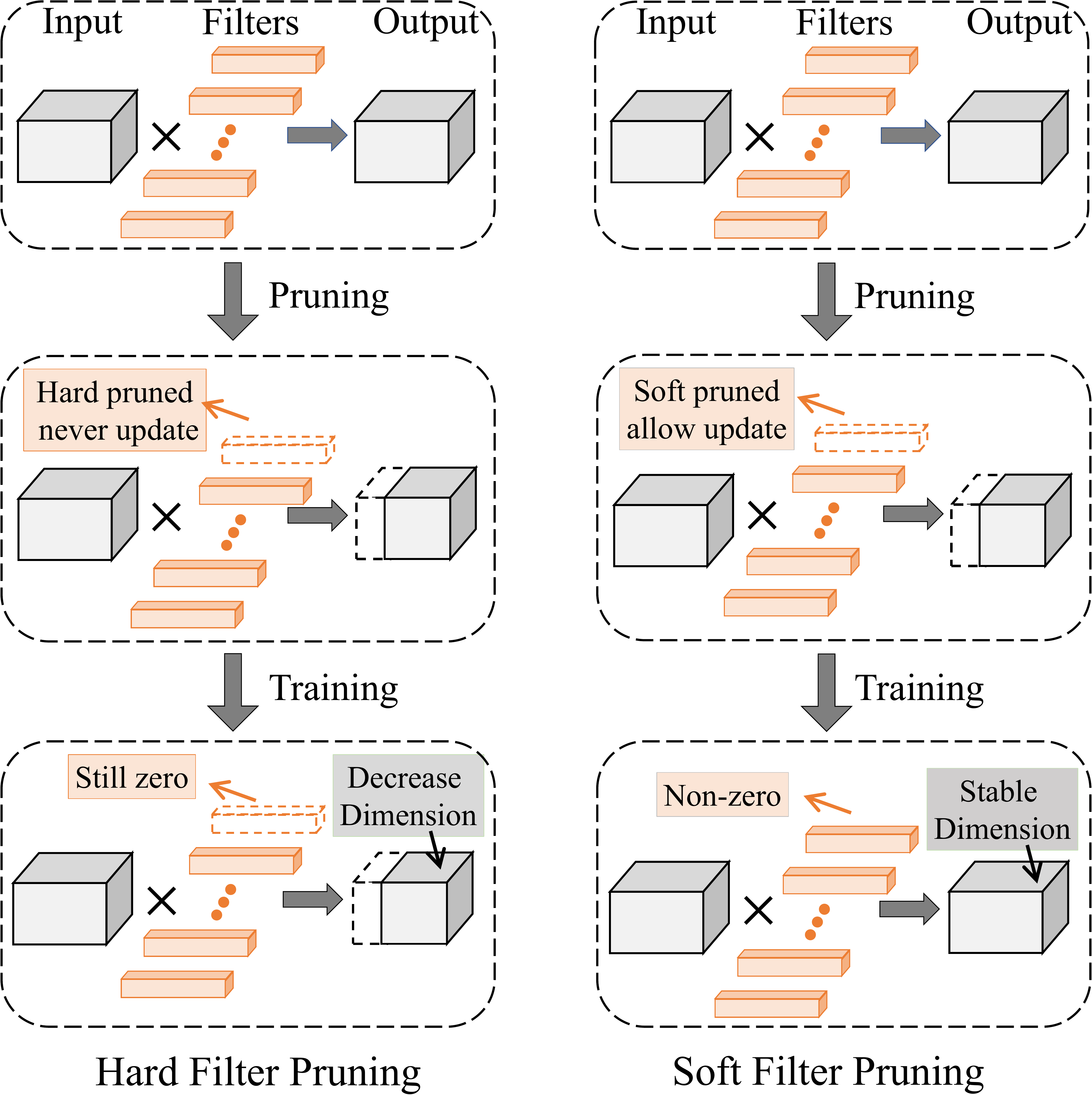} 
\par\end{centering}
\caption{
Hard filter pruning \emph{v.s.} soft filter pruning.
We mark the pruned filter as the orange dashed box.
For the hard filter pruning, the pruned filters are always \textbf{\emph{fixed}} during the whole training procedure.
Therefore, the model capacity is reduced and thus harms the performance because the dashed blue box is useless during training.
On the contrary, our soft pruning method allows the pruned filters to be \textbf{\emph{updated}} during the training procedure.
In this way, the model capacity is recovered from the pruned model and thus leads a better accuracy.
}
\label{fig:hard_soft} 
\end{figure}

Pruning deep CNNs~\cite{hassibi1993second,reed1993pruning,zeng2018accelerating} is an important direction for accelerating the
networks. Recent efforts have been made either on directly deleting some weight values of filters~\cite{han2015learning} (\emph{i.e.}, weight pruning) or totally discarding some filters (\emph{i.e.}, filter pruning)~\cite{li2016pruning,He_2017_ICCV,Luo_2017_ICCV}.
However, weight pruning results in the \emph{unstructured sparsity} of filters. Since the unstructured model cannot leverage the existing high-efficiency BLAS (Basic Linear Algebra Subprograms) libraries, weight pruning is not efficient in saving the memory usage and computational cost.
In contrast, filter pruning enables the model with \emph{structured sparsity}, taking full advantage of BLAS libraries to achieve more efficient memory usage and more realistic acceleration.
Therefore, filter pruning is more favored in accelerating the networks.

Nevertheless, most filter pruning algorithms suffer from two problems: (1) \emph{the model capacity reduction} and (2) \emph{the unrecoverable filter information loss}.
Specifically, as shown in Figure~\ref{fig:hard_soft}, most researchers conduct the ``hard filter pruning (HFP)''~\cite{li2016pruning,He_2017_ICCV,Luo_2017_ICCV}, then the pruned filters are directly deleted and have no possibility to be recovered.
The discarded filters will reduce the optimization space and model capacity, and thus it is unfavorable for the pruned network to learn enough knowledge.
To alleviate this problem, they use pre-training to maintain a good performance, which in turn induces much more training time.
Furthermore, existing methods directly prune a large number of filters, which contain information of training set, at first. This process leads to severe and unrecoverable information loss and thus inevitably degrades the performance.
% \iffalse
% point 2
% (2) \emph{the requirement of pre-trained model}
% Second, these methods always have a three-stage pipeline, \emph{i.e.}, training, pruning, and retraining. In other words, pre-training the model is necessary to maintain good performance.
% Therefore, they often require much more training time and has lower training efficiency than the traditional training schema.
% \fi

To solve the above two problems, we propose ASFP, which prunes the convolutional filters dynamically.
Particularly, before the first training epoch, the filters with small $\ell_{2}$-norm are selected and set to zero.
Then we retrain the model, and the previously pruned filters could be updated. Before the next training epoch, we will prune a \emph{new set} of filters with small $\ell_{2}$-norm.
These training processes are continued until converged.
Lastly, some filters with smallest $\ell_{2}$-norm will be selected and pruned without further updating. This soft manner enables the compressed network to have a larger optimization space and model capacity.
Hence it is easier for the model to learn from the training data, and achieve higher accuracy even without the pre-training process.

In addition, we prune the network asymptotically ------ pruning few filters at first, and more filters at later training epochs.
If few filters are pruned at first, little pre-trained information would be lost, so it is easy for the model to recover from pruning.
With the following iterative training and soft pruning, the training set information would be gradually concentrated in some important filters.
At the same time, the training and pruning process would be stable, as the information is lost gradually instead of suddenly.
%less informative filters are becoming discriminative and easy to be recognized, which would 

We highlight the following three contributions of ASFP:

\noindent
(1) We propose a soft manner to allow the previous pruned filters to be reconstructed during training. This soft manner could significantly maintain the model capacity, which enables the network to be trained and pruned simultaneously from scratch.

\noindent
(2) To avoid severe information loss, we propose to asymptotically prune the filters, which makes the subsequent training and pruning process more stable.

\noindent
(3) Experiments on CIFAR-10 and ImageNet demonstrate the effectiveness and efficiency of the proposed ASFP.

\section{Related Work}

CNN accelerating methods can be roughly divided into four categories, namely, \emph{matrix decomposition}, \emph{low-precision weights}, \emph{weight pruning}, and \emph{filter pruning}.

%Essentially, the work of this paper is based on the idea of pruning techniques; and the approaches of matrix decomposition and low-precision weights are orthogonal but potentially useful here – it may be still worth simplifying the weight matrix after pruning filters, which would be taken as future work.

\subsubsection{Matrix Decomposition}
To reduce the computation costs of the convolutional layers, previous work propose to representing the weight matrix of the convolutional network as a low-rank product of two smaller matrices~\cite{jaderberg2014speeding,zhang2016accelerating,zhang2015efficient,tai2015convolutional,park2016faster,chien2018tensor}. Then the calculation of production of one large matrix turns to the production of two smaller matrices. 
However, the computational cost of tensor decomposition operation is expensive, which is not friendly to train deep CNNs. Besides, there exists an increasing usage of $1\times1$ convolution kernel in some recent neural networks, such as the bottleneck block structure of ResNet~\cite{he2016deep}, cases where it is difficult to apply matrix decomposition.

\subsubsection{Low Precision}
Some other researchers focus on low-precision implementation to compress and accelerate CNN models~\cite{han2015deep,zhu2016trained,zhou2017incremental,hubara2016binarized,rastegari2016xnor}. 
Zhou \emph{et~al.}~\cite{zhu2016trained} propose trained ternary quantization to reduce the precision of weights in neural networks to ternary values. The authors of~\cite{zhou2017incremental} present incremental network quantization, targeting to convert pre-trained full-precision CNN model into a low-precision version efficiently.
%whose weights are constrained to be either power of two or zero.
In this situation, only low-precision weights are stored and used during the inference procedure, with the storage and computation cost being dramatically reduced.

\subsubsection{Weight Pruning}
Recent work~\cite{han2015learning,han2015deep,guo2016dynamic} prunes weights of neural networks.
For example, \cite{han2015learning} proposed an iterative weight pruning method by discarding the small weights whose values are below the threshold.
%\cite{guo2016dynamic} proposed the dynamic network surgery to reduce the training iteration while maintaining a good prediction accuracy.
%\cite{guo2016dynamic} proposed
%dynamic network surgery which is a network compression method.
\cite{wen2016learning,lebedev2016fast} leveraged the sparsity property of feature maps or weight parameters to accelerate the CNN models.
% To this end, \cite{wen2016learning} proposed the Structured Sparsity
% Learning (SSL) method to regularize filter, channel, filter shape
% and depth structures. \cite{lebedev2016fast} applied the group-sparsity
% regularization on the loss function to shrink some entire groups of
% weights towards zeros.
However, weight pruning always leads to unstructured models, so the model cannot leverage the existing efficient BLAS libraries in practice. Therefore, it is difficult for weight pruning to achieve realistic speedup.
Meanwhile, Bayesian methods~\cite{ullrich2017soft} are also applied to network pruning.
%~\cite{ullrich2017soft} extends the soft weight sharing to obtain a sparse and compressed network.
%~\cite{louizos2017bayesian} uses hierarchical priors to prune nodes and utilizes the posterior uncertainties to encode the weights.
%~\cite{molchanov2017variational} uses variational inference to learn the dropout rate which can then be used to prune the network.
However, these methods are evaluated on rather small datasets such as MNIST~\cite{lecun1998gradient} and CIFAR-10~\cite{krizhevsky2009learning}.

\subsubsection{Filter Pruning} Pruning the filters~\cite{li2016pruning,Liu_2017_ICCV,He_2017_ICCV,Luo_2017_ICCV} leads to the removal of the
corresponding feature maps, thus not only reducing the storage usage on devices but also decreasing the memory footprint consumption. 
Considering whether to utilize the training data to determine the pruned filters, the filter pruning methods are roughly divided into two categories, data dependent and data independent filter pruning. The latter method is more efficient than the former since training data may not be available during the pruning process.

\textbf{Data Dependent Filter Pruning.}
Some approaches~\cite{Liu_2017_ICCV,Luo_2017_ICCV,He_2017_ICCV,molchanov2016pruning,dubey2018coreset,yu2018nisp,zhuang2018discrimination,he2018adc,huang2018learning} utilize the training data to determine the pruned filters.
The authors of \cite{dubey2018coreset} minimize the reconstruction error of activation maps to obtain a decomposition of convolutional layers.
Luo \emph{et~al.}~\cite{Luo_2017_ICCV} adopt the statistics information from the next layer to guide the importance evaluation of filters.

%\cite{suau2018principal} proposes an inherently data-driven method which use Principal Component Analysis (PCA) to specify the proportion of the energy that should be preserved.
%\cite{wang2018exploring} applies subspace clustering to feature maps to eliminate the redundancy in convolutional filters.
%\cite{Liu_2017_ICCV} imposes sparsity regularization on the scaling factors of the network.
%\cite{He_2017_ICCV} utilizes the LASSO regression to select channels.
%\cite{yu2018nisp} proposes to minimize the reconstruction error of important responses in the “final response layer”, and derives a closed-form solution to it for pruning neurons in earlier layers.

\textbf{Data Independent Filter Pruning.}
Concurrently with our work, some data independent filter pruning strategies~\cite{li2016pruning,he2018soft,ye2018rethinking,he2019filter} have been explored. 
Li \emph{et~al.}~\cite{li2016pruning} explore the sensitivity of layers for filter pruning and utilize a $\ell_{1}$-norm criterion to prune unimportant filters.
Ye \emph{et~al.}~\cite{ye2018rethinking} prune models by enforcing sparsity on the scaling parameters of batch normalization layers.
%\cite{he2018soft} proposes to select filters with a $\ell_{2}$-norm criterion and prune those selected filters in a soft manner.
%\cite{zhuo2018scsp} uses spectral clustering on filters to select unimportant ones.
However, for all these filter pruning methods, the representative capacity of the neural network after pruning is seriously affected by smaller optimization space. Besides, the information loss at the beginning is significant and unrecoverable.

%\cite{Liu_2017_ICCV} introduces $\ell_{1}$ regularization on the scaling factors in batch normalization (BN) layers as a penalty term, and prune channel with small scaling factors in BN layers. \cite{molchanov2016pruning} proposes a Taylor expansion based pruning criterion to approximate the change in the cost function induced by pruning. \cite{He_2017_ICCV} proposes a LASSO-based channel selection strategy, and a least-square reconstruction algorithm to prune filters.

\iffalse
\subsection{Discussion.}
To the best of our knowledge, there is only one approach that uses a soft manner to prune weights~\cite{guo2016dynamic}.
We would like to highlight our advantages compared to this approach as below:
1) Our pruning algorithm focuses on the filter pruning, but they focus on the weight pruning.
As discussed above, weight pruning approaches lack practical implementations to achieve realistic acceleration.
2) \cite{guo2016dynamic} paid more attention to the model compression, whereas our approach can achieve both compression and acceleration of the model.
3) Extensive experiments have been conducted to validate the effectiveness of our proposed approach both on large-scale datasets and the state-of-the-art CNN models.
In contrast, \cite{guo2016dynamic} only had the experiments on AlexNet which is more redundant than the advanced models, such as ResNet.
\fi

%%%%%%%%%%%%%%%%%%%%%%%%%%%%%%%%%%%%%%%%%%%%%%%%%%%%%%%%%%%%%%%%%%%%%%%%%%%%%%%%%%%%%%%%%%%%%%%%%%%

%%%%%%%%%%%%%%%%%%%%%%%%%%%%%%%%%%%%%%%%%%%%%%%%%%%%%%%%%%%%%%%%%%%%%%%%%%%%%%%%%%%%%%%%%%%%%%%%%%%
\section{Methodology}
\label{Soft Filter Pruning}

%This section firstly gives the problem setup in Sec. \ref{Preliminary};then the details of SFP algorithm is fully developed in Sec. \ref{Soft Filter Pruning}. Finally, some theoretical analysis is given in Sec. \ref{Calculation Reduction After Pruning}.
% In this section, we will give a comprehensive introduction to our SFP and
% present its implementation details.
%%%%%%%%%%%%%%%%%%%%%%%%%%%%%%%%%%%%%%%%%%%%%%%%%%%%%%%%%%%%%%%%%%%%%%%%%%%%%%%%%%%%%%%%%%%%%%%%%%%

\subsection{Preliminary \label{Preliminary}}

We formally introduce the symbol and notations in this section.
The deep CNN network can be parameterized by $\{\mathbf{W}^{(i)}\in\mathbb{R}^{N_{i+1}\times N_{i}\times K\times K},1\leq i\leq L\}$
\footnote{Fully-connected layers can be viewed as convolutional layers with $k=1$}.
$\mathbf{W}^{(i)}$ denotes a matrix of connection weights in the $i_{th}$ layer.
$N_{i}$ denotes the number of input channels for the $i_{th}$ convolution layer.
$L$ denotes the number of layers.
The shapes of input tensor $\mathbf{U}$ and output tensor $\mathbf{V}$ are $N_{i} \times H_{i}\times W_{i} $ and $N_{i+1} \times H_{i+1}\times W_{i+1} $, respectively.
The convolutional operation of the $i_{th}$ layer can be written as:
{
\begin{align}
\label{eq:1}
\mathbf{V}_{i,j}=\mathcal{F}_{i,j}\ast\mathbf{U} ~,~ \mathbf{for}~1 \leq j \leq N_{i+1},
\end{align}
}
\noindent where $\mathcal{F}_{i,j} \in \mathbb{R}^{N_{i}\times K\times K}$ represents the $j_{th}$ filter of the $i_{th}$ layer, and $\mathbf{V}_{i,j}$ represents the $j_{th}$ output feature map of the $i_{th}$ layer.
$\mathbf{W}^{(i)}$ consists of $\{\mathcal{F}_{i,j}, 1 \leq j \leq N_{i+1}\}$.
% and $\mathcal{F}_{i,j}^{k}$ represents the $k$-th kernel of $\mathcal{F}_{i,j}$.
% 
% which is composed by kernels
% $\mathcal{K}$. Applying $N_{i+1}$ filters
% $\mathcal{F}_{i,j}\in\mathbb{R}^{N_{i}\times k\times k}$ on the $N_{i}$
% input channels generates $N_{i+1}$ feature maps.

Pruning filters can remove the output feature maps.
In this way, the computational cost of the neural network will reduce remarkably.
Let us assume the pruning rate is $P_{i}$ for the $i_{th}$ layer.
The number of filters of this layer will be reduced from $N_{i+1}$ to $N_{i+1}(1-P_{i})$, thereby the size of the output tensor $\mathbf{V}_{i,j}$ can be reduced to ${N_{i+1}(1-P_{i})\times H_{i+1} \times W_{i+1}}$.
%As the output tensor of $i_{th}$ layer is the input tensor of $i+1_{th}$ layer, we can reduce the input size of $i_{th}$ layer to achieve a higher acceleration ratio.

%%%%%%%%%%%%%%%%%%%%%%%%%%%%%%%%%%%%%%%%%%%%%%%%%%%%%%%%%%%%%%%%%%%%%%%%%%%%%%%%%%%%%%%%%%%%%%%%%%%

\begin{figure}[!t]
\begin{centering}
\includegraphics[width=0.4\textwidth]{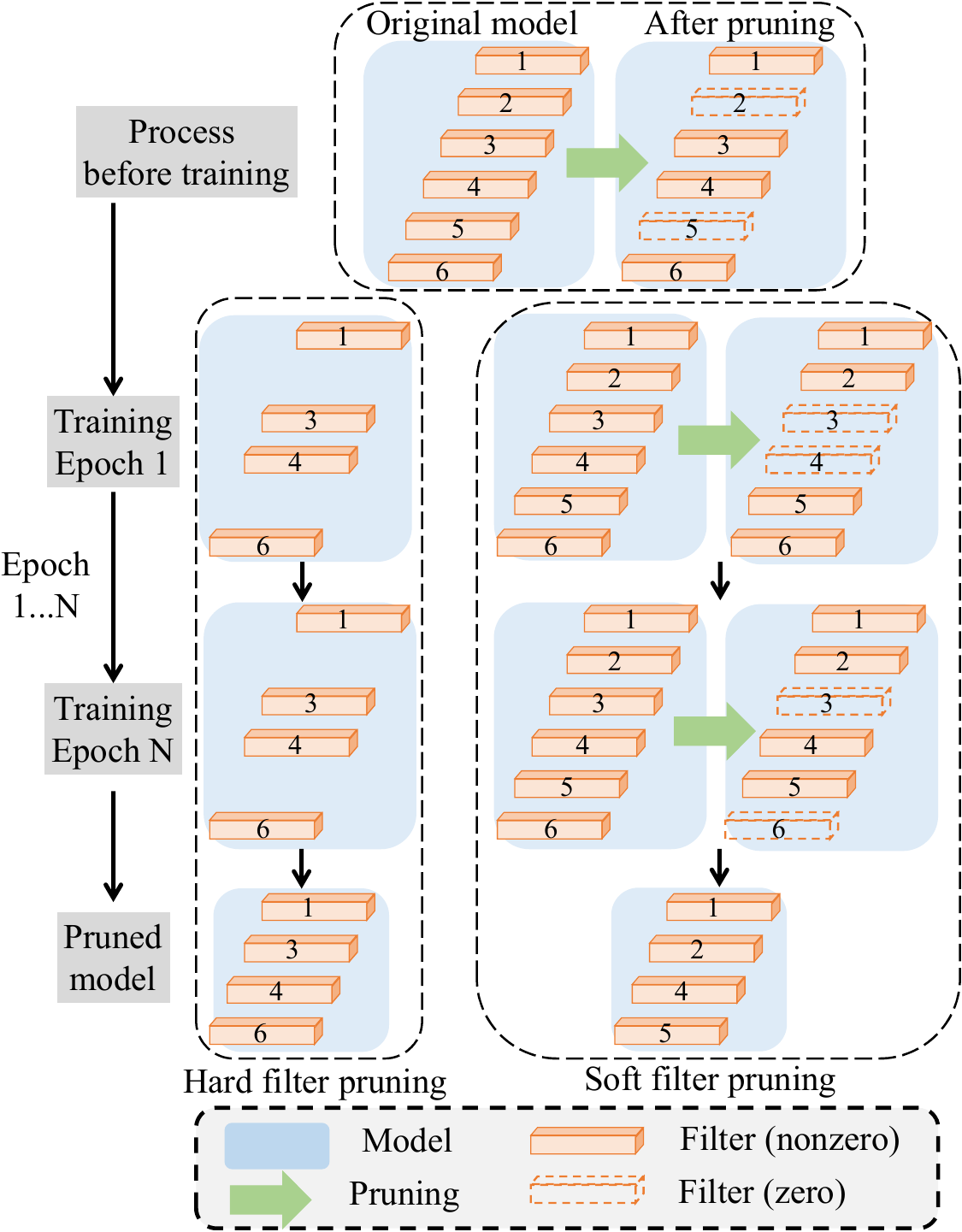}
\par\end{centering}
\caption{Pruning and training schedule of HFP and SFP. 
%The blue area represents the model and the green arrow is the pruning operation. The solid and dashed orange box indicates the nonzero and zero value filter, respectively.
Before training, we first select some filters with pre-defined importance evaluations. HFP directly deletes these filters before training, while for SFP, those are set to zero and kept.
During training (epoch 1 to N), the model size is smaller than the original one for HFP.
While for SFP, the zero value filters (filter 2 and 5) become non-zero after training epoch 1. Then we evaluate the importance of filters again and prune filter 3 and 4. The model size would not be reduced but be the same as the original one.
When training is finished, the final pruned model is the model at epoch \textit{N} for HFP. While for SFP, we delete the zero value filters (filter 3 and 6) at epoch \textit{N} to get the final pruned model.
}
\label{fig:training_compare} 
\end{figure}

\begin{figure*}[!t]
\begin{centering}
\includegraphics[width=0.8\textwidth]{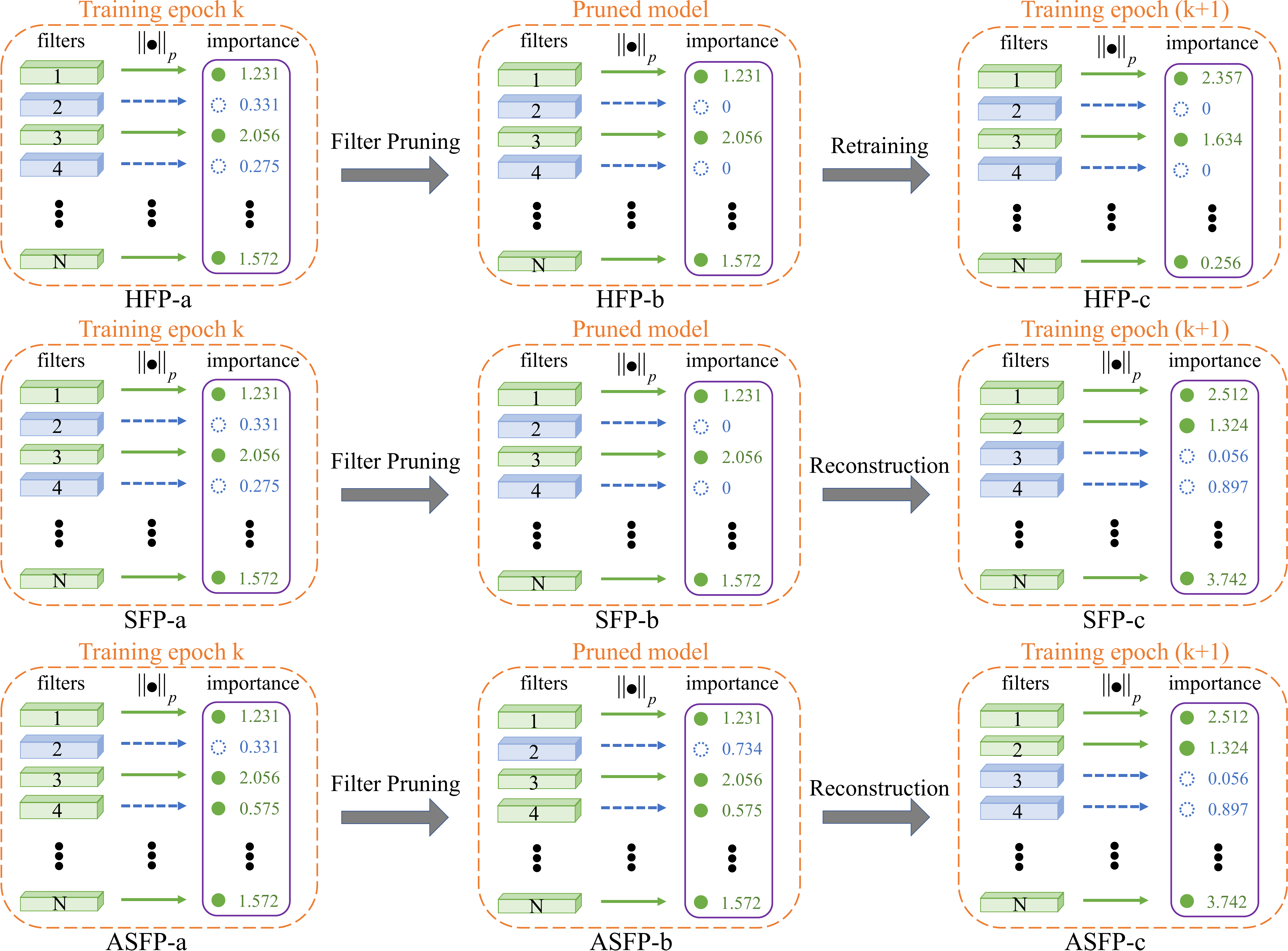} 
\par\end{centering}
\caption{Overview of HFP (first row), SFP (second row) and ASFP (third row).
\textbf{(a)}: filter instantiations before pruning.~\textbf{(b)}: filter instantiations after pruning.~\textbf{(c)}: filter instantiations after reconstruction. 
The filters are ranked by their $\ell_{p}$-norms and the small ones (purple rectangles) are selected to be pruned.
%For HFP, after filter pruning, the pruned filter would have no chance to come back during the retraining process. This is represented by that HFP-b and HFP-c have the same blue filters with zero $\ell_{p}$-norm. For SFP, after filter pruning, the model undergoes a reconstruction process where the pruned filters are capable of being reconstructed (\emph{i.e.} updated from zeros) by the forward-backward process. The number of pruned filters is the same during all training epochs, which is showed by that the number of blue filters in SFP-a is the same as the number of blue filters in SFP-c. While for ASFP, the number of pruned filters is increasing asymptotically during training epochs, which is represented by that ASFP-c has a larger number of blue filters than ASFP-a.
}
\label{fig:soft_pruning} 
\end{figure*}

%%%%%%%%%%%%%%%%%%%%%%%%%%%%%%%%%%%%%%%%%%%%%%%%%%%%%%%%%%%%%%%%%%%%%%%%%%%%%%%%%%%%%%%%%%%%%%%%%%%

\subsection{Pruning with Hard Manner}

%%%%%%%%%%%%%%%%%%%%%%%%%%%%%%%%%%%%%%%%%%%%%%%%%%%%%%%%%%%%%%%%%%%%%%%%%%%%%%%%%%%%%%%%%%%%%%%%%%%
% \begin{algorithm}[t]
% \caption{Algorithm Description of HFP} \label{alg:HFP}
% \begin{algorithmic}
% \Require training data: $\mathbf{X}$, pruning rate: $P_{i}$, the model with parameters $\mathbf{W} = \{\mathbf{W} ^{(i)}, 0\leq i \leq L\}$.
% \State Initialize the model parameter $\mathbf{W}$
% 	\For{$i=1$; $i \leq L $; $i++$}
% 	\State Calculate the $\ell_2$-norm for each filter $\|\mathcal{F}_{i,j}\|_2, 1 \leq j \leq N_{i+1}$
% 	\State Zeroize $N_{i+1}P_i$ filters by $\ell_2$-norm filter selection
% 	\State Compact model $\mathbf{W} ^{'}\in\mathbb{R}^{N_{i+1}(1-P_{i})\times N_{i}(1-P_{i-1})\times K\times K}$
% \For{$epoch=1$; $epoch \leq epoch_{max}$; $epoch++$}
% 	\State Update the model parameter $\mathbf{W}^{'}$ based on $\mathbf{X}$
% 	\EndFor
% \EndFor
% \Ensure The compact model and its parameters $\mathbf{W} ^{*}$
% %Model $\mathbf{W}^{*}\in \mathbb{R}^{N_{i+1}(1-P_{i})\times N_{i}(1-P_{i-1})\times K\times K}$
% \end{algorithmic} 
% \end{algorithm}
%%%%%%%%%%%%%%%%%%%%%%%%%%%%%%%%%%%%%%%%%%%%%%%%%%%%%%%%%%%%%%%%%%%%%%%%%%%%%%%%%%%%%%%%%%%%%%%%%%%

%Most previous filter pruning works \cite{li2016pruning,Liu_2017_ICCV,He_2017_ICCV,Luo_2017_ICCV} prune in a hard manner. We call them the hard filter pruning (HFP).

Given a dataset $\gD = \{(\rvx_i, \rvy_i)\}_{i=1}^n$ and a desired sparsity level $\kappa$ (\ie, the number of remaining non-zero filters), HFP can formulated as:
\begin{align}\label{eq:nn}
\min_{\gF} \ell(\gF;\gD) &= \min_{\gF} \frac{1}{n} \sum_{i=1}^n \ell(\gF; (\rvx_i, \rvy_i))\
,\\\nonumber 
\text{s.t.}  \quad\gF  &\in \R^{N \times K\times K},  \quad \etN (\gF) \le \kappa\ .
\end{align}

Here, $\ell(\cdot)$ is the standard loss function (\eg, cross-entropy loss), $\gF$ is the set of filters of the neural network, and $\etN$ is the cardinality of the filter set.
Typically, HFP firstly prunes filters of a single layer of a pre-trained model and fine-tune the pruned model to complement the performance degradation.
Then they prune the next layer and fine-tune the model again until the last layer is pruned. Once the filters are pruned, HFP will not update these filters again.

The full training schedule of HFP is shown in the first column of Figure~\ref{fig:training_compare}, and the detailed pruning process of HFP is shown in the first row of Figure~\ref{fig:soft_pruning}. First, some of the filters with small $\ell_{p}$-norm (filter 2 and 4 in HFP-a, marked in blue) are selected and pruned. After retraining, these pruned filters are not able to be updated again thus the $\ell_{p}$-norm of these filters would be zero during all the training epochs (filter 2 and 4 in HFP-b).
Meanwhile, the remaining filters (filter 1, 3 and N in HFP-c, marked in green) might be updated to another value after retraining to make up for the performance degradation due to pruning. After several epochs of retraining to converge the model, a compact model is obtained to accelerate the inference. 
%%%%%%%%%%%%%%%%%%%%%%%%%%%%%%%%%%%%%%%%%%%%%%%%%%%%%%%%%%%%%%%%%%%%%%%%%%%%%%%%%%%%%%%%%%%%%%%%%%%

\subsection{Pruning with Soft Manner}

%%%%%%%%%%%%%%%%%%%%%%%%%%%%%%%%%%%%%%%%%%%%%%%%%%%%%%%%%%%%%%%%%%%%%%%%%%%%%%%%%%%%%%%%%%%%%%%%%%%
SFP can dynamically remove the filters in a soft manner. Specifically, the key is to keep updating the pruned filters in the training stage. The full training schedule of SFP is shown in the second column of Figure~\ref{fig:training_compare}.
Such an updating manner brings several benefits. It not only keeps
the model capacity of the pruned models as the original models but also avoids the greedy layer by layer pruning procedure and enables pruning~\emph{all} convolutional layers at the same time. 
For SFP, the constrain in the Eq.~\ref{eq:nn} changes to:
\begin{align}\label{eq:nn2}
 \quad \|\gF\|_0 &\le \kappa, \quad \etN (\gF) = N_{i+1}.
\end{align}
Here, $\|\cdot\|_0$ is the standard $L_0$ norm. After soft pruning, the number of filters $\etN (\gF)$ is still the same as that of the original model ($N_{i+1}$).

The second row of Figure~\ref{fig:soft_pruning} explains the detailed process of SFP.
First, the $\ell_{2}$-norms of all filters are computed for each weighted layer and used as our filter selection criterion.
Second, some filters with a small $\ell_{p}$-norm (filter 2 and 4 in SFP-a, marked in blue) are selected, and we prune those filters by setting the corresponding filter weights as zero (filter 2 and 4 in SFP-b).
Then we retrain the model.
As we allow the pruned filters to be updated during retraining, the pruned filters become nonzero again (filter 2 and 4 in SFP-c) due to back-propagation.
After iterations of pruning and reconstruction to converge the model, we delete the unimportant filters and obtain a compact and efficient model.

\subsection{Asymptotic Soft Filter Pruning (ASFP)}

It is known that all the filters of the pre-trained model have the information of the training set. 
Therefore, pruning those informative filters would cause information loss. This situation is especially difficult for pruning the models that pre-trained on large datasets, or pruning a large number of filters of small models, as the lost information is rather massive.
Therefore, we propose to pruning the neural network asymptotically. 

Specifically, we use a small pruning rate at early epochs, and gradually increase the pruning rate later, until we reach the goal pruning rate $P_i$ at the last retraining epoch.
The optimization problem of ASFP:
\begin{align}\label{eq:nn3}
\min_{\gF} \ell (\gF;\gD) &= \min_{\gF} \frac{1}{n} \sum_{i=1}^n \ell(\gF; (\rvx_i, \rvy_i))\
,\\\nonumber 
\text{s.t.} \quad \|\gF\|_0 &\le \kappa_{epoch}, \quad \etN (\gF) = N_{i+1}.
\end{align}
\noindent Here, $\kappa_{epoch}$ means the sparsity level changes with the training epoch. 
Comparing Eq.~\ref{eq:nn2} and Eq.~\ref{eq:nn3}, the difference between SFP and ASFP is whether the pruning rate is changing with epochs. 
For SFP, the $\kappa$ in Eq.~\ref{eq:nn2} is a pre-defined constant regarding the number of remaining non-zero filters, so it would not change during the training process.
In contrast, $\kappa_{epoch}$ in Eq.~\ref{eq:nn3} is a function of $epoch$ and would change with the epoch number of training.
Therefore, SFP is a special case of ASFP when the pruning rate $P_i$ during all training epochs are the same.
The details of ASFP is illustratively explained in Algorithm~\ref{alg:PFP}.

\subsubsection{Asymptotic Filter Selection}
We use the $\ell_{p}$-norm to evaluate the importance of each filter as Eq.~\ref{eq:p-norm}.
In general, the convolutional results of the filters with the smaller $\ell_{p}$-norm would lead to relatively lower activation values, and thus have a less numerical impact on the final prediction of deep CNN models. 
In term of this understanding, such filters of small $\ell_{p}$-norm will be given higher priority of being pruned than those of higher $\ell_{p}$-norm:
{
\begin{align}
\label{eq:p-norm}
\|\mathcal{F}_{i,j}\|_{p}=\sqrt[p]{\sum_{n=1}^{N_{i}}\sum_{k_{1}=1}^{K}\sum_{k_{2}=1}^{K}\left|\mathcal{F}_{i,j} (n,k_{1},k_{2})\right|^{p}} .
\end{align}
}
\noindent In practice, we use the $\ell_{2}$-norm based on the empirical analysis. 

Different from SFP~\cite{he2018soft} that the pruning rate equals the goal pruning rate $P_i^{goal}$ during retraining, we use different pruning rate $P_{i}^{'}$ at every epoch. 
The definition of $P_{i}^{'}$ is list as follows:
{
\begin{align}
\label{eq:progress_rate}
P_{i}^{'} = \mathcal{H}(P_{i}^{goal},D,P_{i}^{min},epoch) ,
\end{align}
}\noindent where $P_{i}^{goal}$ represents the goal pruning rate for the $i_{th}$ layer, $D$ and $P_{i}^{min}$ are pre-defined parameter which will be explicitly explained later. As exponential parameter decay is widely used in optimization~\cite{kingma2014adam} to achieve a stable result, we change the pruning rate exponentially. The equation is as follows:
\begin{align}
\label{eq:progress_exp}
P_{i}^{'} = a \times e^{-k\times epoch} + b ,
\end{align}
\noindent In order to solve three parameters $a,k,b$ of the above exponential equation, three points consists of (epoch, $P_{i}^{'}$) pair are needed. Certainly, the first point is (0, $P_{i}^{min}$), which means the pruning rate is $P_{i}^{min}$ for the first training epoch. In addition, to achieve the goal pruning rate $P_{i}$ at the final retraining epoch, the point ($epoch_{max}$, $P_{i}^{goal}$) is essential. Now we have to define the third point ($epoch_{max} \times D$, $3P_{i}^{goal} / {4}$), to solve the equation. This means that when the epoch number is $epoch_{max} \times D$, the pruning rate increase to $3 / {4}$ of the goal pruning rate $P_{i}^{goal}$. 

%%%%%%%%%%%%%%%%%%%%%%%%%%%%%%%%%%%%%%%%%%%%%%%%%%%%%%%%%%%%%%%%%%%%%%%%%%%%%%%%%%%%%%%%%%%%%%%%%%%

\begin{algorithm}[t]
\SetKwInOut{Input}{Input}
\SetKwInOut{Output}{Output}
\SetKw{KwBy}{by}

%\underline{function ASFP} $(a,b)$\;
\Input{training data $\mathbf{X}$ and training epoch $epoch_{max}$\;
the model with parameters $\mathbf{W} = \{\mathbf{W} ^{(i)}, 0\leq i \leq L\}$\;
pruning rate $P_{i}^{goal}$, $P_{i}^{min}$ and pruning rate decay $D$\;
}
\Output{The compact model and its parameters $\mathbf{W} ^{*}$}
  \For{$epoch\gets 1$ \KwTo $epoch_{max}$ \KwBy $1$}{
    Update the model parameter $\mathbf{W}$ based on $\mathbf{X}$\;
    \For{$i\gets 1$ \KwTo $L$ \KwBy $1$}{
      Calculate $P_{i}^{'}$ with Eq.~\ref{eq:progress_rate}\;
      Get the number of filters in $i$ layer $N_{i+1}$ with $\mathbf{W}$\;
  %    Calculate the $\ell_2$-norm for each filter $\|\mathcal{F}_{i,j}\|_2, 1 \leq j \leq N_{i+1}$ \;
      Zeroize $N_{i+1}P_{i}^{'}$ filters with small $\ell_2$-norm\;
   }
}
Obtain the compact model with parameters $\mathbf{W} ^{*}$ from $\mathbf{W}$

\caption{Algorithm Description of ASFP}\label{alg:PFP}
\end{algorithm}

%%%%%%%%%%%%%%%%%%%%%%%%%%%%%%%%%%%%%%%%%%%%%%%%%%%%%%%%%%%%%%%%%%%%%%%%%%%%%%%%%%%%%%%%%%%%%%%%%%%

% When $p=2$, we get the $\ell_{2}$-norm of filter. We use
% $\ell_{2}$-norm as our selection criteria of filter. Some other magnitude
% based criteria have also been proposed to evaluate the importance
% of each filter in previous literature such as $\ell_{1}$-norm of
% the filters~\cite{li2016pruning}. After comparing these two criteria,
% we find $\ell_{2}$-norm works better than $\ell_{1}$-norm, so we
% choose the $\ell_{2}$-norm as our selection criteria. The detailed
% experiment is shown in section~\ref{1-norm and 2-norm}.
% 
% We use the $\ell_{2}$-norm of filters to measure the importance of filters.
% Specifically, and the detailed explanation is given in section~\ref{Selection of Filter Within A layer}.
% For $N_{i+1}$ filters within \textit{i}th layer, and the pruning rate is $P_{i}$,
% then $N_{i+1}\times P_{i}$ filters which has the lowest $\|\mathcal{F}_{i,j}\|_{2}$ value is to be pruned in this layer.
% As shown in Figure~\ref{fig:soft_pruning},
% the blue filters indicate the filters that have the smallest $\|\mathcal{F}_{i,j}\|_{2}$,
% and these filters should be pruned.

\subsubsection{Filter Pruning}
We set the value of selected $N_{i+1}P_{i}^{'}$ filters to zero (see the filter pruning step in Figure~\ref{fig:soft_pruning}).
This can temporarily eliminate their contribution to the network output. We prune \emph{all} the weighted layers at the same time.
In this way, we can prune each filter in parallel, which would cost negligible computation time.
In contrast, previous methods~\cite{Luo_2017_ICCV,He_2017_ICCV} always conduct a greedy layer by layer pruning and retraining,  which would cost more computation time, especially when the model depth increases.
Moreover, we use the \emph{same} pruning rate for \emph{all} the weighted layers, $P_{i}^{goal}=P$.
Therefore, we only require one hyper-parameter $P$ to balance acceleration and accuracy.
This can avoid the inconvenient hyper-parameter search or the complicated sensitivity analysis shown in~\cite{li2016pruning}.  

%Why can SPF prune all layers by the same pruning rate at the same time?
%Intuitively, this is due to the large model capacity of the model trained by SPF.
%As we allow the pruned filters to be updated, the model has the large model capacity and becomes more flexible and thus can well balance the contribution of each filter to the final prediction.
%Therefore, we may have a better performance than the previous greedy method. Besides, less pruning and reconstruction cycles are needed if we prune all layers together.
%Furthermore, we just have one hyper-parameter which is directly related to the acceleration, that is, the pruning rate $P_{i}=P$.

\subsubsection{Reconstruction}
After the pruning step, we train the network for one epoch to reconstruct the pruned filters.
In order to keep the representative capacity and the high performance of the model, those pruned filters are updated to non-zero by back-propagation, as shown in Figure~\ref{fig:soft_pruning}.
In this way, the pruned model has the same capacity as the original model during the training.
In contrast, hard filter pruning leads to a decrease of feature maps, so the model capacity is reduced and the performance is influenced.
With large model capacity, we could integrate the pruning step into the normal training schema, that is, training and pruning the model synchronously.
%Previous pruning methods usually require a pre-trained model and then fine-tune it.

\subsubsection{Obtaining Compact Model}
ASFP iterates over the filter selection, filter pruning, and reconstruction steps.
After model convergence, we can obtain a sparse model containing many ``zero filters".
The features maps corresponding to those ``zero filters" will always be zero during the inference procedure.
Therefore, there will be no influence to remove these filters as well as the corresponding feature maps.
%Specifically, for the pruning rate $P_{i}$ in the $i_{th}$ layer, only $N_{i+1}(1-P_{i})$ filters are non-zero and have an effect on the final prediction. Consider pruning the previous layer, the input channel of $i_{th}$ layer is changed from $N_{i}$ to $N_{i}(1-P_{i-1})$. We can thus re-build the $i_{th}$ layer into a smaller one.Finally, a compact model $\{\mathbf{W^{*}}^{(i)}\in\mathbb{R}^{N_{i+1}(1-P_{i})\times N_{i}(1-P_{i-1})\times K\times K}\}$ is obtained.

\subsection{Pruning Strategy for Convolutional Network}
\label{Pruning strategy}

\begin{figure}[!t]
\begin{centering}
\includegraphics[width=0.45\textwidth]{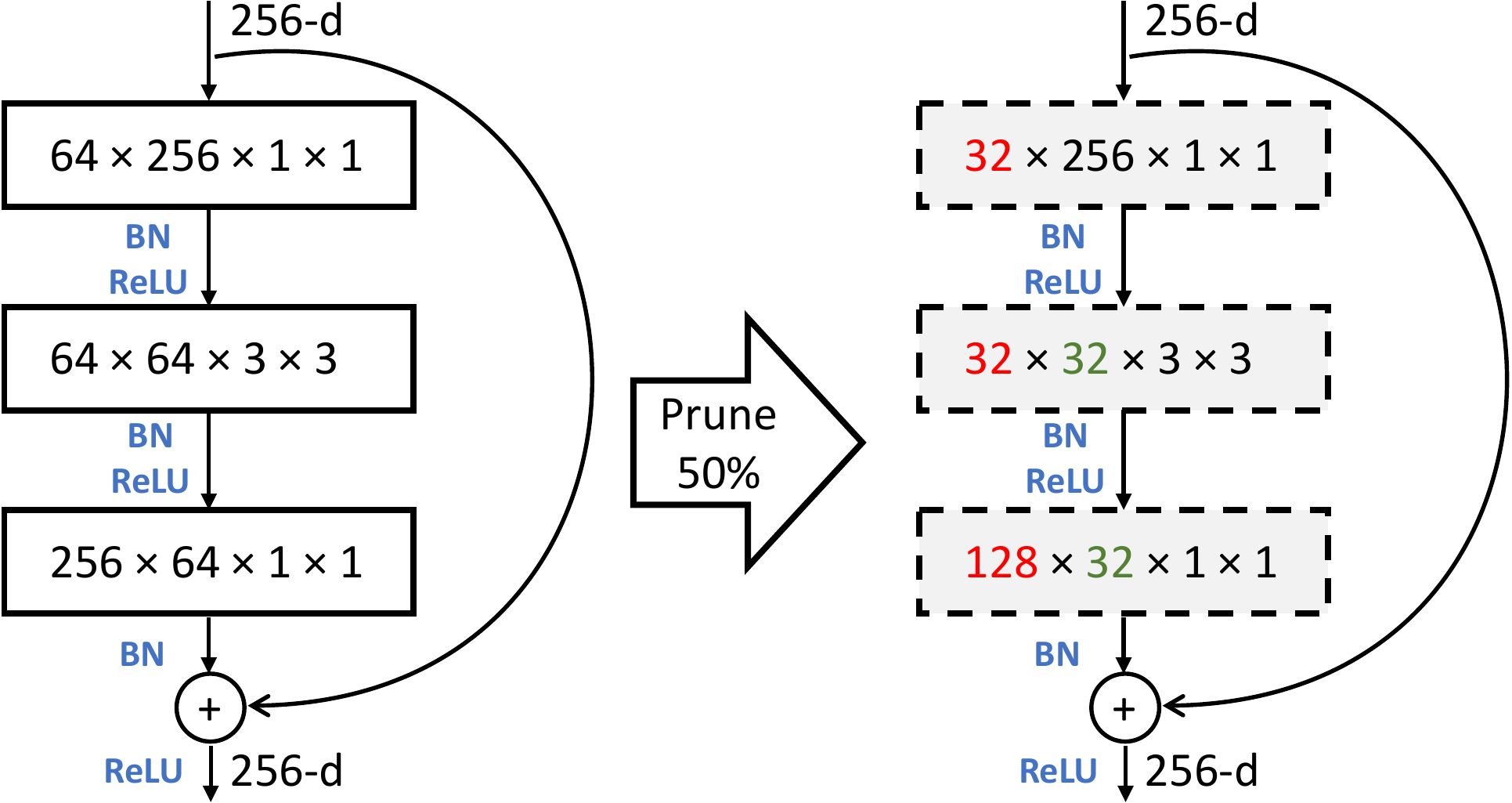}
\par\end{centering}
\caption{Pruning residual block with pruning rate 50\%. Red and green number means the remaining output channel number and input channel number after pruning, respectively. ``BN'' and ``ReLU'' represents the batch norm layer and non-linear layer, respectively. 
}
\label{fig:structure} 
\end{figure}

Pruning the traditional convolutional architectures~\cite{krizhevsky2012imagenet,simonyan2014very} is easy to understand, but it is elusive for pruning recent structural variants, such as ResNet~\cite{he2016deep}. 
In Figure~\ref{fig:structure}, pruning the residual blocks is illustrated.
If we set the pruning rate as 50\%, the output dimension of three layers would decrease by 50\% (the red numbers). The input dimension of the last two layers would change according to the output dimension of the previous layer (the green numbers).
What we should especially care about is the element-wise additive of the residual connections and the output of convolutional layers.
In Figure~\ref{fig:structure}, the channel number of the residual connections and the output of convolutional layers is 256 and 128, respectively.
The element-wise additive should change to:
\begin{equation}
\label{eq:filter}
\rmO_{index}=\left\{
\begin{aligned}
& \rmR_{index} +  \rmC_{index} \quad if \quad index \in \rmI \\
&   \rmR_{index} \quad \quad \quad \quad \quad else\\
\end{aligned}
\right.
\end{equation}
\noindent where $\rmI$ is the index of 128 remaining channels, $\rmC$ is the convolutional output after the batch norm layer (128 channels), $\rmR$ is the residual output (256 channels), and $\rmO$ is the output of the whole residual block.

\subsection{Computation Complexity Analysis}
\label{Calculation Reduction After Pruning}

\subsubsection{Theoretical Speedup Analysis} Suppose the filter pruning rate of
the $i_{th}$ layer is $P_{i}$, which means the $N_{i+1}\times P_{i}$
filters are set to zero and pruned from the layer, and the other $N_{i+1}\times(1-P_{i})$
filters remain unchanged. Besides, suppose the size of the input and output
feature map of $i_{th}$ layer is $H_{i}\times W_{i}$ and $H_{i+1}\times W_{i+1}$.
Then after filter pruning, the dimension of useful output feature
map of the $i_{th}$ layer decreases from $N_{i+1}\times H_{i+1}\times W_{i+1}$
to $N_{i+1}(1-P_{i})\times H_{i+1}\times W_{i+1}$. Note that
the output of $i_{th}$ layer is the input of ${(i+1)}_{th}$ layer.
And we further prunes the ${(i+1)}_{th}$ layer with a filter pruning rate
$P_{i+1}$, then the calculation of ${(i+1)}_{th}$ layer is decrease
from $N_{i+2}\times N_{i+1}\times k^{2}\times H_{i+2}\times W_{i+2}$
to $N_{i+2}(1-P_{i+1})\times N_{i+1}(1-P_{i})\times k^{2}\times H_{i+2}\times W_{i+2}$. In other words, a proportion of $1-(1-P_{i+1})\times(1-P_{i})$ of the original
calculation is reduced, which will make the inference procedure of CNN models much faster.

\subsubsection{Realistic Speedup Analysis} In theoretical speedup analysis, other operations such as batch normalization and pooling are negligible compared to the convolution operations. Therefore, we consider the FLOPs (Floating-point operations per second) of convolution operations for computation complexity comparison, which is commonly used in the previous work~\cite{li2016pruning,Luo_2017_ICCV}.
In the real scenario, reduced FLOPs cannot bring the same level of realistic speedup because non-tensor layers (\textit{e.g.}, batch normalization and pooling layers) also need the inference time on GPU~\cite{Luo_2017_ICCV}. In addition, the limitation of IO delay, buffer switch, and efficiency of BLAS libraries also lead to the wide gap between theoretical and realistic speedup ratio.
We compare the theoretical and realistic speedup in Section~\ref{section:ILSVRC}.

\section{Experiment}

%%%%%%%%%%%%%%%%%%%%%%%%%%%%%%%%%%%%%%%%%%%%%%%%%%%%%%%%%%%%%%%%%%%%%%%%%%%%%%%%%%%%%%%%%%%%%%%%%%%

%{\renewcommand{\arraystretch}{1.0}
\begin{table*}[th] 
\centering  
\setlength{\tabcolsep}{1.1em}
\caption{
Overall performance of pruning ResNet on CIFAR-10.
} 	
\begin{tabular}{c|c c c c c c c c}  		
\hline 		
Depth & Method &Pre-train? &Baseline Accu. (\%)  &Accelerated Accu. (\%)  &Accu. Drop (\%) & FLOPs & Pruned FLOPs(\%)       
\\ \hline \hline 			 	        

%\multirow{4}{*}{20}       
%&MIL~\cite{Dong_2017_CVPR} &\xmark       & 91.53        & 91.43 &0.10 &	3.20E7	  & 20.3	 \\          
%& SFP~\cite{he2018soft}(10\%) &\xmark       & \textbf{92.20} ($\pm$0.18)       &  \textbf{92.24} ($\pm$0.33)  &\textbf{-0.04} & 3.44E7 	&15.2\\         
%& SFP~\cite{he2018soft}(20\%) &\xmark       & \textbf{92.20} ($\pm$0.18)       & 91.20 ($\pm$0.30)  & 1.00	& 2.87E7 	&29.3  \\         
%& SFP~\cite{he2018soft}(30\%) &\xmark       & \textbf{92.20} ($\pm$0.18)       &  90.83 ($\pm$0.31) & 1.37&  \textbf{2.43E7} &\textbf{42.2}  \\ \hline  \hline 	         

%\multirow{4}{*}{32} &MIL~\cite{Dong_2017_CVPR} &\xmark       &92.33         & 90.74 &1.59 &	4.70E7	  & 31.2	 \\          
%& SFP~\cite{he2018soft}(10\%)   &\xmark    & \textbf{92.63} ($\pm$0.70)       &   \textbf{93.22} ($\pm$0.09) & \textbf{-0.59}&  5.86E7 &14.9  \\         
%& SFP~\cite{he2018soft}(20\%)   &\xmark    & \textbf{92.63} ($\pm$0.70)       &  90.63 ($\pm$0.37) & 0.00 &  4.90E7 &28.8  \\   
%& SFP~\cite{he2018soft}(30\%)   &\xmark    & \textbf{92.63} ($\pm$0.70)       &  90.08 ($\pm$0.08) & 0.55 &  \textbf{4.03E7} &\textbf{41.5} \\\hline  \hline 	

\multirow{8}{*}{56} 		
&PFEC~\cite{li2016pruning} &\xmark     & 93.04                   & 91.31  &1.75 &9.09E7  &27.6		 \\         

&CP~\cite{He_2017_ICCV} &\xmark       & 92.80                      & 90.90 &1.90 &	-	  & 50.0	 \\    

%& SFP~\cite{he2018soft}(10\%)   &\xmark 	&\textbf{93.59} ($\pm$0.58)   & \textbf{93.89} ($\pm$0.19) & \textbf{-0.30}&  1.070E8 &14.7 \\        
%& SFP~\cite{he2018soft}(20\%)   &\xmark 	&\textbf{93.59} ($\pm$0.58)   & 93.47 ($\pm$0.24) & 0.12 &8.98E7 &28.4\\       
%& SFP~\cite{he2018soft}(30\%)   &\xmark 	&\textbf{93.59} ($\pm$0.58)  & 93.10 ($\pm$0.20) & \textbf{0.49}  &  {7.40E7} &41.1 \\  
& SFP~\cite{he2018soft}    &\xmark 	&\textbf{93.59} ($\pm$0.58)  & 92.26 ($\pm$0.31) & 1.33   &  \textbf{5.94E7} &\textbf{52.6} \\        
%& ASFP(30\%)   &\xmark 	&\textbf{93.59} ($\pm$0.58)   & 93.05 ($\pm$0.19) & 0.54 & {7.40E7} &41.1  \\
& ASFP~(40\%)   &\xmark 	&\textbf{93.59} ($\pm$0.58)   & 92.44 ($\pm$0.07) &\textbf{1.15}  & \textbf{5.94E7} &\textbf{52.6} \\
\cdashline{2-8} 

&PFEC~\cite{li2016pruning}  &\cmark     & 93.04          & 93.06 &\textbf{-0.02} &9.09E7 & 27.6 \\	
&CP~\cite{He_2017_ICCV} &\cmark       & 92.80                   & 91.80 &1.00 &	-	 &	50.0 	 \\      
%& SFP~\cite{he2018soft}(30\%)   &\cmark 	&\textbf{93.59} ($\pm$0.58)   & 93.78 ($\pm$0.22) & -0.19 & {7.40E7} &41.1  \\
& SFP~\cite{he2018soft}    &\cmark 	&\textbf{93.59} ($\pm$0.58)   & \textbf{93.35} ($\pm$0.31) &0.24   & \textbf{5.94E7} &\textbf{52.6} \\    
%& ASFP(30\%)   &\cmark 	&\textbf{93.59} ($\pm$0.58)   & 93.25 ($\pm$0.23) & 0.34 & {7.40E7} &41.1  \\
& ASFP~(40\%)   &\cmark 	&\textbf{93.59} ($\pm$0.58)   & 93.12 ($\pm$0.20) &0.47   & \textbf{5.94E7} &\textbf{52.6} \\ \hline \hline

\multirow{10}{*}{110}          		&PFEC~\cite{li2016pruning} &\xmark  &93.53     & 92.94      &  0.61              & 1.55E8 	&38.6 	 \\          
&MIL~\cite{Dong_2017_CVPR}  &\xmark       & 93.63         & 93.44 &0.19 &	-	  & 34.2 	 \\          
%& SFP~\cite{he2018soft}(10\%)   &\xmark  & \textbf{93.68} ($\pm$0.32)	& 93.83 ($\pm$0.19)  & -0.15 &  2.16E8 &14.6 \\  	

%& SFP~\cite{he2018soft}(20\%)   &\xmark & \textbf{93.68} ($\pm$0.32)	&\textbf{93.93} ($\pm$0.41) & \textbf{-0.25 }&  1.82E8 &28.2 \\  		
%& SFP~\cite{he2018soft}(30\%)   &\xmark & \textbf{93.68} ($\pm$0.32) 	& 93.38 ($\pm$0.30) & 0.30& {1.50E8} &{40.8}   \\  		

& SFP~\cite{he2018soft}   &\xmark & \textbf{93.68} ($\pm$0.32) 	& 92.62 ($\pm$0.60) & 1.04& \textbf{1.21E8} &\textbf{52.3}   \\  	

%& ASFP(10\%)   &\xmark  & \textbf{93.68} ($\pm$0.32)	& 93.70 ($\pm$0.55)  & -0.02 &  2.16E8 &14.6 \\  	
& ASFP~(20\%)   &\xmark & \textbf{93.68} ($\pm$0.32)	&\textbf{93.94} ($\pm$0.56) & \textbf{-0.24}&  1.82E8 &28.2 \\  		
%& ASFP(30\%)   &\xmark & \textbf{93.68} ($\pm$0.32) 	& 93.29 ($\pm$0.45) & 0.39& {1.50E8} &{40.8}   \\  		
& ASFP~(40\%)   &\xmark & \textbf{93.68} ($\pm$0.32) 	& 93.20 ($\pm$0.10) & 0.48& \textbf{1.21E8} &\textbf{52.3}   \\  		
\cdashline{2-8} 

&PFEC~\cite{li2016pruning} &\cmark    & 93.53    & 93.30  &0.20 	&1.55E8 	&38.6\\ 
& SFP~\cite{he2018soft}    &\cmark & \textbf{93.68} ($\pm$0.32) 	& \textbf{93.86} ($\pm$0.21) & \textbf{-0.18} & {1.50E8} &{40.8}   \\    
& SFP~\cite{he2018soft}   &\cmark & \textbf{93.68} ($\pm$0.32) 	& 92.90 ($\pm$0.18) & 0.78& \textbf{1.21E8} &\textbf{52.3}   \\  	
%& ASFP(10\%)   &\cmark  & \textbf{93.68} ($\pm$0.32)	& 93.46 ($\pm$0.12)  & -0.22 &  2.16E8 &14.6 \\  	
%& ASFP(20\%)   &\cmark & \textbf{93.68} ($\pm$0.32)	&93.38 ($\pm$0.06) & {-0.30 }&  1.82E8 &28.2 \\  		
& ASFP~(30\%)   &\cmark & \textbf{93.68} ($\pm$0.32) 	& 93.37 ($\pm$0.12) & 0.31& {1.50E8} &{40.8}   \\  		
& ASFP~(40\%)   &\cmark & \textbf{93.68} ($\pm$0.32) 	& 93.10 ($\pm$0.06) & 0.58& \textbf{1.21E8} &\textbf{52.3}   \\  		
\hline  	
\end{tabular}  	

\label{table:cifar10_accuracy} 
\end{table*}  

%%%%%%%%%%%%%%%%%%%%%%%%%%%%%%%%%%%%%%%%%%%%%%%%%%%%%%%%%%%%%%%%%%%%%%%%%%%%%%%%%%%%%%%%%%%%%%%%%%%

%%%%%%%%%%%%%%%%%%%%%%%%%%%%%%%%%%%%%%%%%%%%%%%%%%%%%%%%%%%%%%%%%%%%%%%%%%%%%%%%%%%%%%%%%%%%%%%%%%%
\label{Experiment}
%In this section, we conduct experiments on several benchmark datasets to validate the effectiveness of our acceleration method.

%%%%%%%%%%%%%%%%%%%%%%%%%%%%%%%%%%%%%%%%%%%%%%%%%%%%%%%%%%%%%%%%%%%%%%%%%%%%%%%%%%%%%%%%%%%%%%%%%%%

\subsection{Benchmark Datasets and Experimental Setting}

\textbf{Dataset.}
Our method is evaluated on two benchmarks:  CIFAR-10~\cite{krizhevsky2009learning} and ILSVRC-2012~\cite{russakovsky2015imagenet}. 
CIFAR-10 contains 50,000 training images and 10,000 test images, which are categorized into ten classes.
ILSVRC-2012 is a large-scale dataset containing 1.28 million training images and 50k validation images of 1,000 classes.

\textbf{Architecture.}
As discussed in \cite{Luo_2017_ICCV,He_2017_ICCV,Dong_2017_CVPR}, multiple-branch ResNet~\cite{he2016deep} is less redundant than VGGNet~\cite{simonyan2014very}, so it is more difficult to accelerate ResNet.
Therefore, we focus on pruning the challenging ResNet model.
To validate our method on the single-branch network, we also prune the VGGNet following~\cite{li2016pruning}.

\textbf{Training setting.} 
In the CIFAR-10 experiments, we use the default parameter setting in~\cite{he2016identity} and follow the training schedule in~\cite{zagoruyko2016wide}. For CIFAR-10 dataset, we test our ASFP on ResNet-56 and 110. 
On ILSVRC-2012, we follow the same parameter settings as~\cite{he2016deep,he2016identity}.
The data argumentation strategies are the same as PyTorch implementation~\cite{paszke2017automatic}. We test ASFP on ResNet-18, 34, 50 and we use the pruning rate 30\% for all the models.
We also analyze the difference between pruning the pre-trained model and scratch model.
For pruning the model from scratch, We use the normal training schedule without additional fine-tuning process. For pruning the pre-trained model, we reduce the learning rate to one-tenth of the original learning rate. 
To conduct a fair comparison of pruning from scratch and pre-trained models, we use the same training epochs to train/fine-tune the network. The previous work~\cite{li2016pruning} uses fewer epochs to fine-tune the pruned model, but it converges too early and harms the accuracy, as shown in section ~\ref{sec:vggnet}.

\textbf{Pruning setting.}
For VGGNet on CIFAR-10, we use the same pruning rate as~\cite{li2016pruning}.
For experiments on ResNet, we follow~\cite{he2018soft} and prune \emph{all} the convolutional layers with the \emph{same} pruning rate at the same time. 
We do not prune the projection shortcuts for simplification, which only need negligible time and do not affect the overall cost.
Therefore, only one hyper-parameter, the pruning rate $P_{i}=P$ is used to balance acceleration and accuracy.

\begin{figure}[t]
\center
\includegraphics[width=0.75\linewidth]{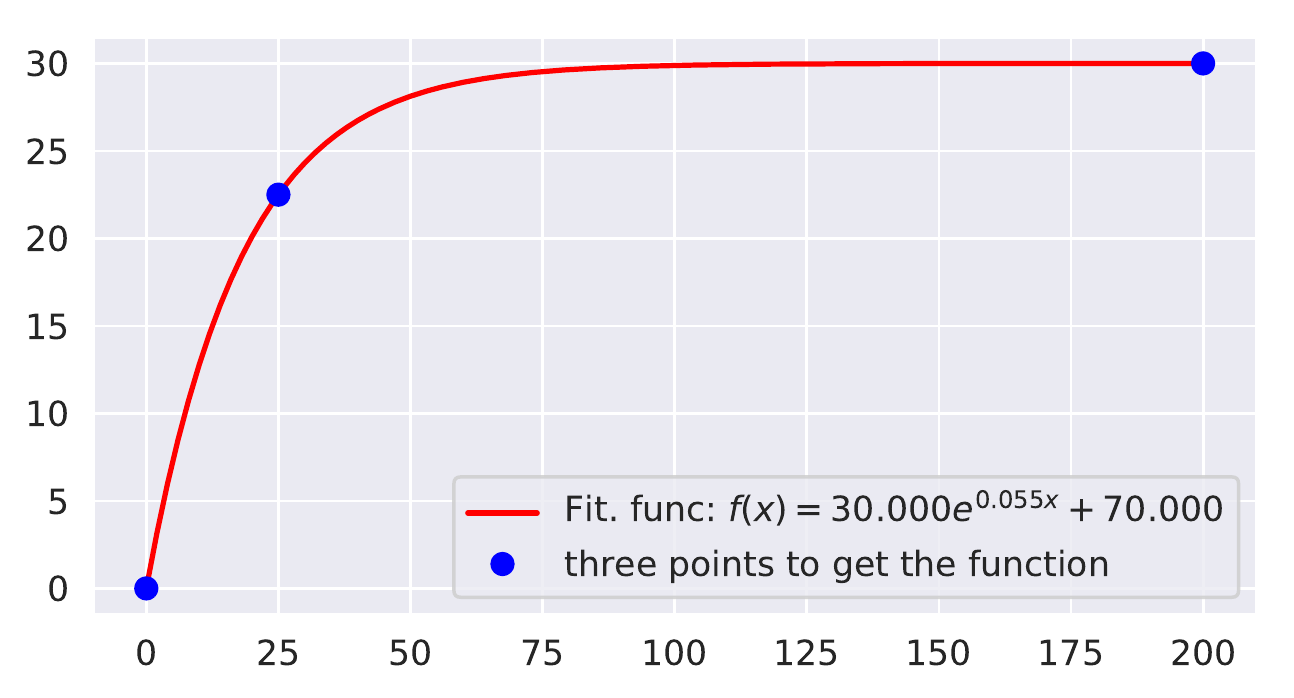}
\caption{
Asymptotically changed pruning rate when the goal pruning rate is 30\%. Three blue points are the three pairs to generate the exponential function of pruning rate (the solid curve).
}
\label{fig:exp_rate}
\vspace{-2mm}
\end{figure}

To asymptotically change the pruning rate, we set the parameters in Eq.~\ref{eq:progress_rate} as $D=1/8$ and $P_{i}^{min}=0$. This setting is denoted as ASFP-P0.\footnote{For the following sections, the ``ASFP'' (without suffix) means ``ASFP-P0'' if not particularly indicated.}
For example, if we use the goal pruning rate $P_{i}^{goal} = 30\%$, then the pruning rate curve according to epoch is shown in Figure~\ref{fig:exp_rate}.
If we set $P_{i}^{min} = P_{i}^{goal}$, ASFP is same as SFP. 
The pruning operation is conducted at the end of every training epoch.
We run some experiments three times and report the ``mean $\pm$ std''.
The performance is compared with other state-of-the-art acceleration algorithms, \eg, MIL~\cite{Dong_2017_CVPR}, PFEC~\cite{li2016pruning}, CP~\cite{He_2017_ICCV}, ThiNet~\cite{Luo_2017_ICCV}, SFP~\cite{he2018soft}, NISP~\cite{yu2018nisp}. 
We choose to directly cite the numbers from original papers for a fair comparison.

\textbf{Explanation for Tables.}
The results of ResNet are listed in Table~\ref{table:cifar10_accuracy} and Table~\ref{table:imagenet_accuracy}.
In ``Pre-train?'' column, ``Y'' and ``N'' indicate whether to use the pre-trained model as initialization or not.
The ``Accu. Drop'' is the accuracy of the pruned model minus that of the baseline model, so negative number means the accelerated model has higher accuracy than the baseline model.
A smaller number of "Accu. Drop" is better.
For CIFAR-10, we run every experiment for three times to get the mean and standard deviation of the accuracy.
For Imagenet, we just list the one-view accuracy.
%The different baseline accuracies are mainly due to the different data augmentation method. In this condition, we could use the “Accu. Drop.” instead of the “Accelerated Accu.” to evaluate the effectiveness of the method. 

\textbf{Explanation for Baselines.}
The baseline network is the same for different pruning methods, and accuracy numbers are cited from the original paper. The different accuracies are due to different hyper-parameter settings (\emph{e.g.} different data augmentations, different learning rate schedules, \emph{etc.}) and different implementation frameworks (\emph{e.g.} Caffe, TensorFlow and Pytorch).
For example, in Thinet~\cite{Luo_2017_ICCV}, all the images are resized into $256 \times 256$, then center-cropped to $224 \times 224$. However, in CP~\cite{He_2017_ICCV}, the images are resized such that the shorter side equals to 256.
In this case, we choose to use the “Accu. Drop” (in Table~\ref{table:cifar10_accuracy}) rather than the “Accelerated Accu.” (in Table~\ref{table:cifar10_accuracy}) to fairly evaluate the effectiveness of our method.

\textbf{Optimization Time of ASFP.}
In this paper, we care more about the acceleration during the inference time rather than the training time. However, we would like to show that the additional time cost of ASFP is negligible compared to a hard pruning method.
Take the scratch model for an example.
HFP and ASFP both need 200 epochs to training CIFAR-10 from scratch to converge, so the additional time cost of ASFP comes from the operation of pruning (Eq.~\ref{eq:nn3}). 
Two steps are needed in such a process. 1) Obtaining the pruning rate $P_{i}^{'}$. 2) Ranking and pruning the filters. For step one, the exponent pattern of the pruning rate is pre-defined, and it could be directly accessed during training. For step two, after we get the importance scores of the filters, we zeroize the filters with smaller importance scores to conduct the pruning operation.
%For step two, given N filters, we first calculate the norm of filters to get a $1 \times N$ vector, which is the importance scores of the filters. Then we zeroize the filters with smaller importance scores to conduct the pruning operation.
All these steps bring minor computation cost.
For HFP, it takes about 171.01s for training ResNet-110 for one epoch on GTX 1080. For ASFP, the total time for one epoch is 171.45s. 
Therefore, the time difference between soft pruning and hard pruning is negligible.

%%%%%%%%%%%%%%%%%%%%%%%%%%%%%%%%%%%%%%%%%%%%%%%%%%%%%%%%%%%%%%%%%%%%%%%%%%%%%%%%%%%%%%%%%%%%%%%%%%%

\subsection{VGGNet on CIFAR-10}
\label{sec:vggnet}
The result of pruning from scratch and pre-trained VGGNet is shown in Table~\ref{table:vgg}. Not surprisingly, ASFP achieves better performance than~\cite{li2016pruning} in both settings.
With our pruning criterion, we could achieve slightly better accuracy than ~\cite{li2016pruning} when pruning the random initialized VGGNet (93.37\% \textit{vs.} 93.31\%).
In addition, The pruned model without fine-tuning has better performance than~\cite{li2016pruning} (81.66\% \textit{vs.} 77.45\%). 
After fine-tuning 40 epochs, our model achieves similar accuracy with~\cite{li2016pruning}. 
Notably, if more fine-tuning epochs (160) are utilized, the accuracy of~\cite{li2016pruning} is almost unchanged (93.28\% \textit{vs.} 93.22\%), which means their models have no much more capacity to learn. 
On the contrary, our method could achieve much better performance (94.02\% \textit{vs.} 93.28\%) with more fine-tuning epochs, which shows the model capacity of our ASFP is much larger than~\cite{li2016pruning}. 

%------------------------------------------------------

\begin{table}[ht]
\setlength{\tabcolsep}{1em}
\caption{Pruning from scratch and pre-trained VGGNet on CIFAR-10. ``FT" means ``fine-tuning" the pruned model.}
\begin{center}
\begin{tabular}{c|c|c}
\hline
{Setting $\backslash$ Acc (\%)} & PFEC~\cite{li2016pruning} & Ours  \\  \hline
{Baseline }                   & 93.58 ($\pm$0.03)         & 93.58 ($\pm$0.03) \\
Prune from scratch            & 93.31 ($\pm$0.03)         & \textbf{93.37} ($\pm$0.08) \\
Prune from pre-train without FT & 77.45 ($\pm$0.03)         & \textbf{81.66} ($\pm$0.03) \\
FT 40 epochs                  & 93.22 ($\pm$0.03)         & \textbf{93.27} ($\pm$0.08) \\
FT 160 epochs                 & 93.28 ($\pm$0.03)         & \textbf{94.02} ($\pm$0.15) \\\hline 
\end{tabular}
\end{center}

\label{table:vgg}
\end{table}
%------------------------------------------------------

%%%%%%%%%%%%%%%%%%%%%%%%%%%%%%%%%%%%%%%%%%%%%%%%%%%%%%%%%%%%%%%%%%%%%%%%%%%%%%%%%%%%%%%%%%%%%%%%%%%
%\setlength{\tabcolsep}{0.9em} % for the horizontal padding
%{\renewcommand{\arraystretch}{1.3}% for the vertical padding
%\shortstack \cline{2-7} 
%{\renewcommand{\arraystretch}{1.1}
\begin{table*}[ht]  
\setlength{\tabcolsep}{0.7em} 
\centering 
\caption{
Overall performance of pruning ResNet on ImageNet.
} 
\begin{tabular}{c|c c c c c c c c c c} 		
\hline    		
\multirow{2}{*}{Depth}	   & \multirow{2}{*}{Method}  &\multirow{2}{*}{\shortstack {Pre-\\train?}}  &\multirow{2}{*}{\shortstack {Top-1 Accu.\\Baseline(\%)} }  &\multirow{2}{*}{\shortstack {Top-1 Accu.\\Accelerated(\%)} } &\multirow{2}{*}{\shortstack {Top-5 Accu.\\Baseline(\%)} }   &\multirow{2}{*}{\shortstack {Top-5 Accu.\\Accelerated(\%)} }   &\multirow{2}{*}{\shortstack {Top-1 Accu.\\ Drop(\%)} }  &\multirow{2}{*}{\shortstack {Top-5 Accu.\\ Drop(\%)} } & \multirow{2}{*}{\shortstack {Pruned\\FLOPs(\%)}}    \\ \\ \hline                

\multirow{5}{*}{18}
&MIL~\cite{Dong_2017_CVPR}  &\xmark   &69.98 &66.33 & 89.24     & 86.94	&3.65	 &2.30   & 34.6		 \\	       		  
& SFP~\cite{he2018soft}  &\xmark &\textbf{70.23} ($\pm$0.06)	&{67.25} ($\pm$0.13) & \textbf{89.51} ($\pm$0.10)  & {87.76} ($\pm$0.06)   &{2.98}   & {1.75} & \textbf{41.8} \\ 
& ASFP~(30\%) &\xmark &\textbf{70.23} ($\pm$0.06)	& \textbf{67.41}   & \textbf{89.51} ($\pm$0.10)  &  \textbf{87.89}   &\textbf{2.82}   & \textbf{1.62} &\textbf{41.8} \\  

\cdashline{2-10} 
& SFP~\cite{he2018soft}  &\cmark &\textbf{70.23} ($\pm$0.06)	&60.79   & \textbf{89.51} ($\pm$0.10)  &  {83.11}   &{9.44}   & {6.40} & \textbf{41.8} \\

& ASFP~(30\%) &\cmark &\textbf{70.23} ($\pm$0.06)	& \textbf{68.02}   & \textbf{89.51} ($\pm$0.10)  & \textbf {88.19}   &\textbf{2.21}   & \textbf{1.32} & \textbf{41.8} \\  \hline

\multirow{5}{*}{34}	          
%&~\cite{Dong_2017_CVPR} &\xmark   & 73.42          &\textbf{72.99} 	&91.36	 &\textbf{91.19}	&\textbf{0.43} 	&\textbf{0.17} 	& 24.8	\\        
& SFP~\cite{he2018soft} 	&\xmark 	&\textbf{73.92}		&\textbf{71.83}	&\textbf{91.62}   & 90.33   & \textbf{2.09}  & 1.29 & \textbf{41.1}    \\    
& ASFP~(30\%)	&\xmark 	&\textbf{73.92}		& 71.72	&\textbf{91.62}   &\textbf{90.65}     & {2.20}  &\textbf{0.97}   & \textbf{41.1}  \\       
\cdashline{2-10} 

&PFEC~\cite{li2016pruning} &\cmark    & \textbf{73.23}          & 72.17 	&-	 &-	&\textbf{1.06} 	&- 	& 24.2	 \\  			  
& SFP~\cite{he2018soft} 	&\cmark  &\textbf{73.92}		&{72.29}	&\textbf{91.62}   & {90.90}   & 1.63  & {0.72} & \textbf{41.1}    \\  
& ASFP~(30\%)	&\cmark 	&\textbf{73.92}		&\textbf{72.53} 	&\textbf{91.62}   & \textbf{91.04}    & {1.39}  &\textbf{0.58}   & \textbf{41.1}   \\      \hline

\multirow{7}{*}{50}	
& SFP~\cite{he2018soft}  &\xmark  &\textbf{76.15}		&{74.61}		&\textbf{92.87}	  & {92.06} &1.54	 & 0.81 & \textbf{41.8}  	 \\
& ASFP~(30\%) &\xmark  &\textbf{76.15}		&\textbf{74.88}		&\textbf{92.87}	  & \textbf{92.39} &\textbf{1.27} 	 &\textbf{0.48}   &\textbf{41.8}   	 \\

\cdashline{2-10}
&CP~\cite{He_2017_ICCV} &\cmark 	&- 	&-	& 92.20    &90.80  	&- 	& 1.40	 &\textbf{50.0}	 	\\     		  
&ThiNet~\cite{Luo_2017_ICCV}   &\cmark  &72.88 	&72.04  & 91.14   & 90.67              & {0.84} 	&{0.47}	& 36.7  \\  	
&NISP~\cite{yu2018nisp} & \cmark  &-		&-	&-	  & - & -	 & 0.89& 44.0  	 \\

& SFP~\cite{he2018soft}  &\cmark  &\textbf{76.15}		&{62.14}		&\textbf{92.87}	  & {84.60} &14.01	 & 8.27 & 41.8  	 \\

& ASFP~(30\%) &\cmark  &\textbf{76.15}		&\textbf{75.53}		&\textbf{92.87}	  & \textbf{92.73} &\textbf{0.62} 	 &\textbf{0.14}   &41.8    	 \\

\hline  

%\multirow{4}{*}{101}	    	
%& SFP~\cite{he2018soft}(30\%) &\xmark  &\textbf{77.37}		& 77.03 	&\textbf{93.56}	  &  93.46  & 0.34 	 &  0.10  & \textbf{42.2}    \\  
%& SFP~\cite{he2018soft}(30\%) &\cmark  &\textbf{77.37}		&\textbf{77.51}		&\textbf{93.56}	  & \textbf{93.71} &\textbf{-0.14}	 & \textbf{-0.15}& \textbf{42.2}    \\
%& ASFP(30\%) &\xmark  &\textbf{77.37}		&77.01		&\textbf{93.56}	  & {93.43} &{0.36}	 & {0.13}& \textbf{42.2}    \\ 	
%& ASFP(30\%) &\cmark  &\textbf{77.37}		&77.28	&\textbf{93.56}	  & 93.69 &0.09	 &-0.13 & \textbf{42.2}    \\ 		
%\hline    	
\end{tabular} 	

\label{table:imagenet_accuracy}
\end{table*}
%%%%%%%%%%%%%%%%%%%%%%%%%%%%%%%%%%%%%%%%%%%%%%%%%%%%%%%%%%%%%%%%%%%%%%%%%%%%%%%%%%%%%%%%%%%%%%%%

\subsection{ResNet on CIFAR-10}

\begin{figure*}[ht]
\begin{minipage}{1\textwidth}
\subfigure[ResNet-56 from Pre-train.]{
\includegraphics[width=0.32\linewidth]{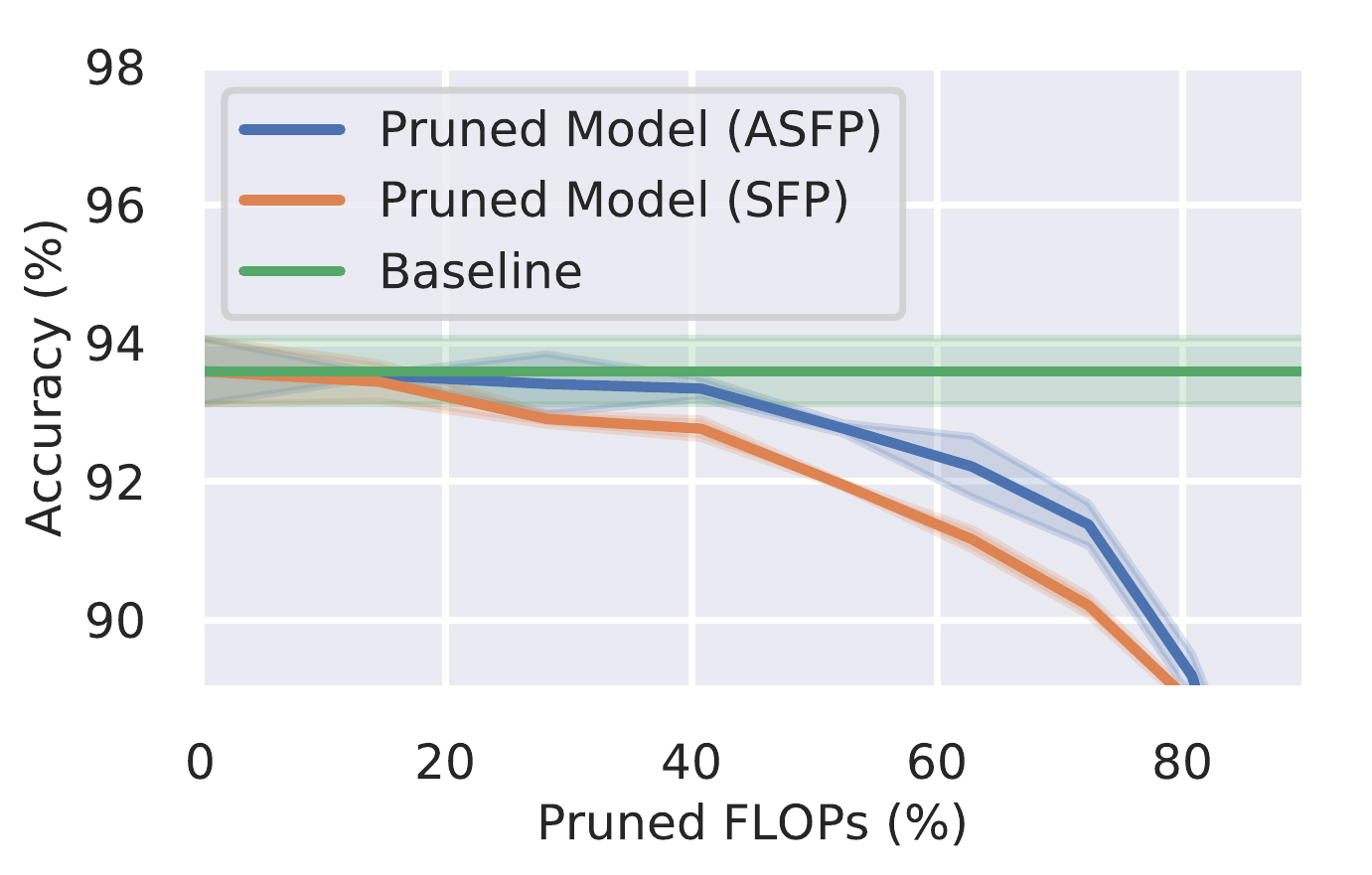}~\label{fig:different_rate_a}
}
\subfigure[ResNet-110 from Pre-train.]{
\includegraphics[width=0.32\linewidth]{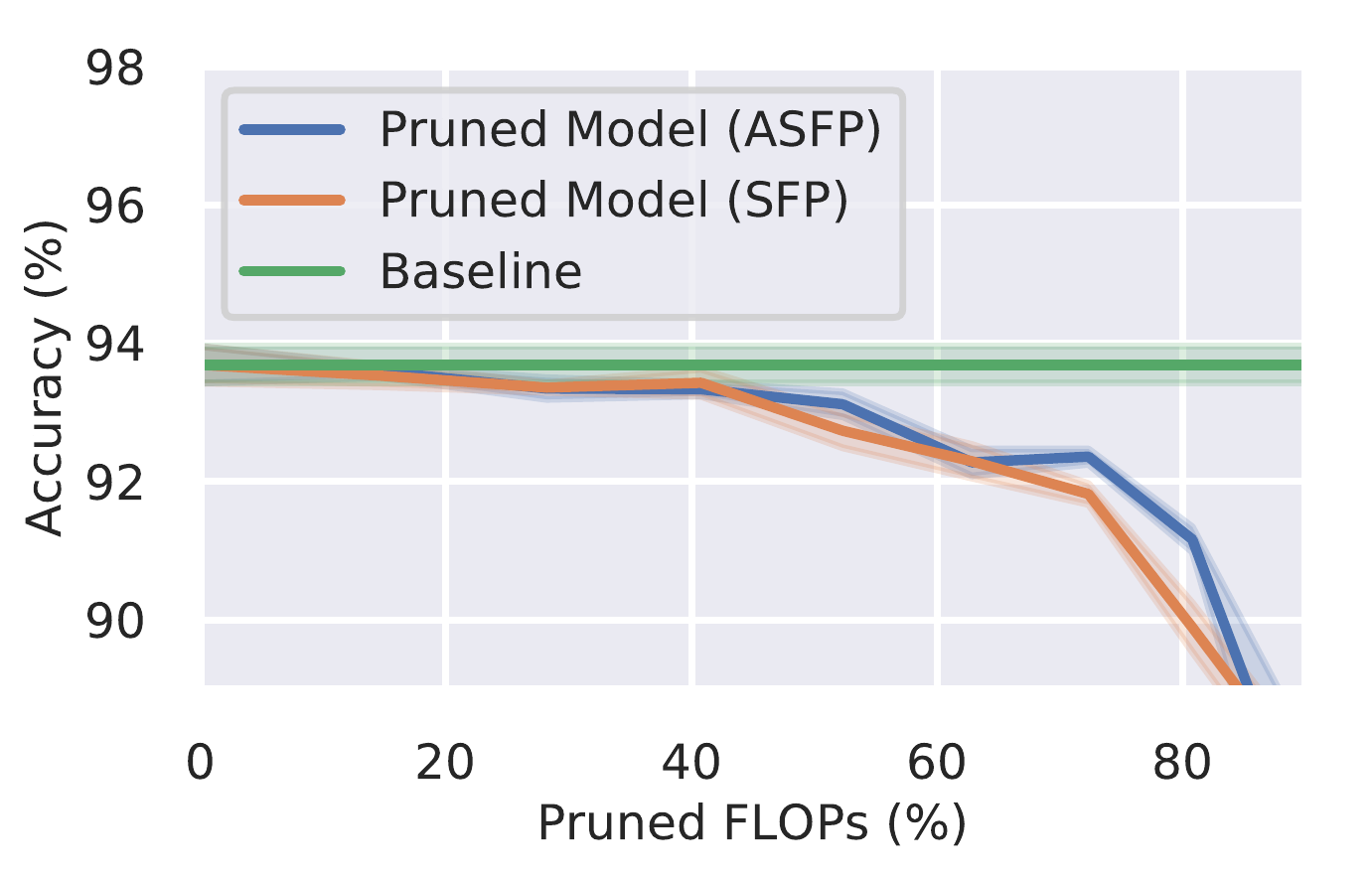}~\label{fig:different_rate_b}
}
\subfigure[ResNet-110 from Scratch.]{
\includegraphics[width=0.32\linewidth]{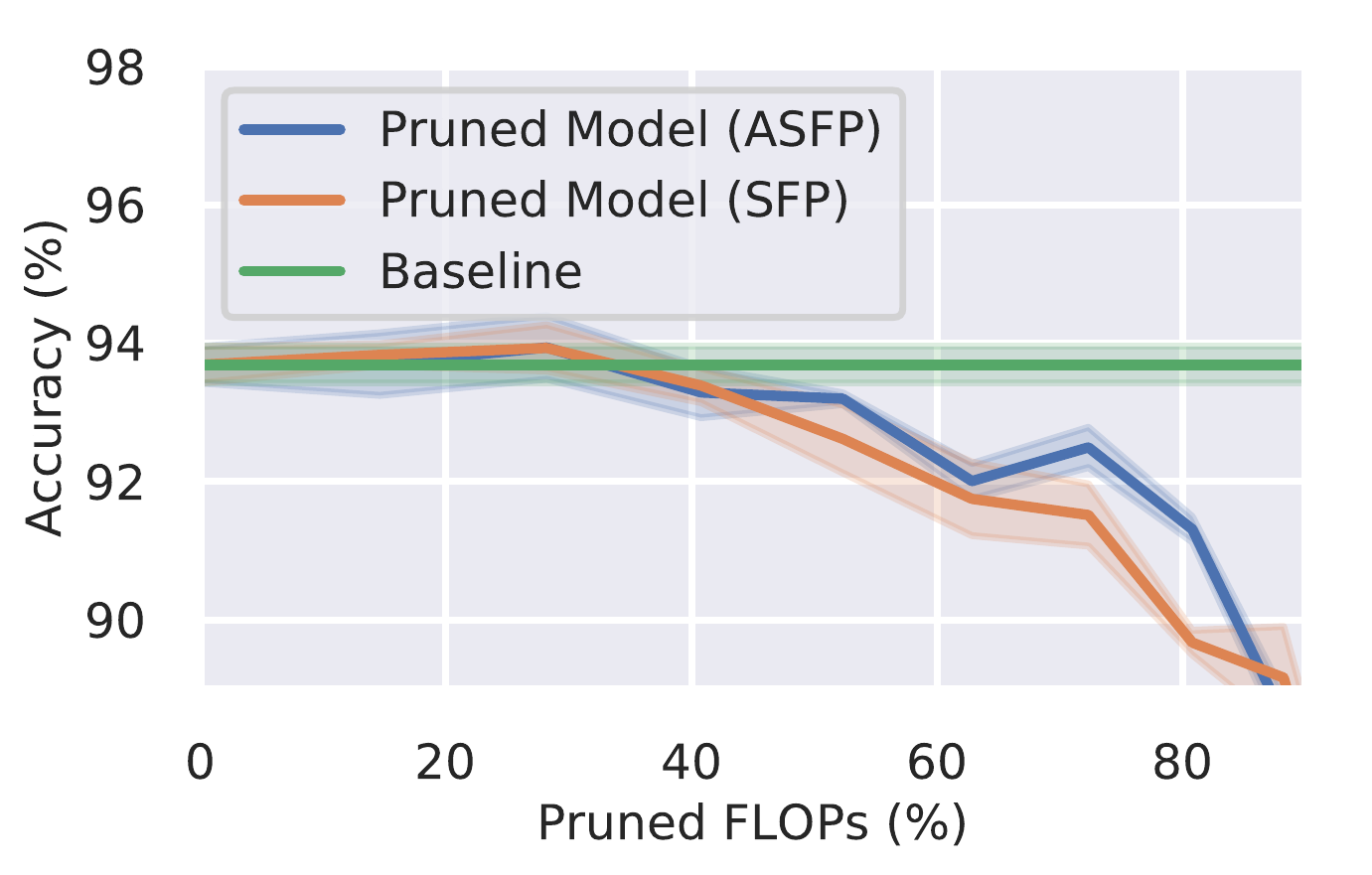}~\label{fig:different_rate_c}
}
\caption{
Model performance regarding different ratio of pruned FLOPs. The green line indicates the model without pruning. The blue and the orange lines represent the model under ASFP and SFP, respectively.}\label{fig:different_rate}
\end{minipage}
%\vspace{-2mm}
\end{figure*}

Table~\ref{table:cifar10_accuracy} shows the results on CIFAR-10.
Our ASFP could achieve a better performance than other state-of-the-art hard filter pruning methods.
For example, PFEC~\cite{li2016pruning} accelerate ResNet-110 by 38.6\% speedup ratio with 0.61\% accuracy drop when pruning the scratch models.
In contrast, our ASFP can accelerate ResNet-110 to 52.3\% speed-up with only 0.48\% accuracy drop.
When pruning the pre-trained ResNet-110, the accuracy drop of our ASFP is smaller than PFEC~\cite{li2016pruning} when pruning the same number of ratio.
When pruning the scratch ResNet-56, we can achieve more acceleration ratio than CP~\cite{He_2017_ICCV} (52.6\% \textit{vs.} 50.0\%) with less accuracy drop (1.15\% \textit{vs.} 1.90\%)
Notably, we can even improve 0.24\% accuracy when pruning 28.2\% FLOPs of scratch ResNet-56.

When the pruning rate is small, we find the performance of SFP~\cite{he2018soft} and ASFP is competitive. But ASFP outperforms SFP when a large portion of FLOPs are pruned. The comprehensive comparison is shown in Figure~\ref{fig:different_rate}.
This is because ASFP is suitable for the situation when a large quantity of the information is removed by pruning.
These results validate the effectiveness of our ASFP algorithm, which can produce a more compressed model with comparable performance to the original model.

%%%%%%%%%%%%%%%%%%%%%%%%%%%%%%%%%%%%%%%%%%%%%%%%%%%%%%%%%%%%%%%%%%%%%%%%%%%%%%%%%%%%%%%%%%%%%%%%%%%
\subsection{ResNet on ILSVRC-2012}
\label{section:ILSVRC}

%%%%%%%%%%%%%%%%%%%%%%%%%%%%%%%%%%%%%%%%%%%%%%%%%%%%%%%%%%%%%%%%%%%%%%%%%%%%%%%%%%%%%%%%%%%%%%%%
%{\renewcommand{\arraystretch}{1.1}
\begin{table}[ht!]
\centering
\caption{
Comparison of the theoretical and realistic speedup.
We only count the time consumption of the forward procedure.
%We use the GTX1080 GPU with a batch size of 64.
}
\begin{tabular}{ c | c | c | c | c}
\hline
  \multirow{2}{*}{Model}	   & \multirow{2}{*}{\shortstack{Baseline\\time (ms)}}  &\multirow{2}{*}{\shortstack {Pruned\\time (ms) }}  &\multirow{2}{*}{\shortstack {Realistic \\Speed-up(\%) } }  &\multirow{2}{*}{\shortstack {Theoretical\\Speed-up(\%)} }\\
&    & & &   \\ \hline
ResNet-18      & 37.10   &  26.97   & 27.4   &  41.8    \\  
ResNet-34      & 63.97   &  45.14   & 29.4   &  41.1    \\  
ResNet-50      & 135.01  &  94.66   & 29.8   &  41.8    \\ 
%ResNet-101     & 219.71  & 148.64   & 32.3   &  42.2    \\ 
\hline
\end{tabular}
\label{table:Comparison_Speed}
\end{table}
%%%%%%%%%%%%%%%%%%%%%%%%%%%%%%%%%%%%%%%%%%%%%%%%%%%%%%%%%%%%%%%%%%%%%%%%%%%%%%%%%%%%%%%%%%%%%%%%%%%
\textbf{Result Explanation.} 
Table~\ref{table:imagenet_accuracy} shows that ASFP outperforms other state-of-the-art methods.
For pruning from a random initialized ResNet-18, our ASFP achieves more inference speedup than MIL~\cite{Dong_2017_CVPR} (41.8\% v.s. 34.6\%), but the top-5 accuracy drop of our pruned model is less than that of their model (1.62\% v.s. 2.30\%).
For pruning pre-trained ResNet-34, our ASFP achieves a much better acceleration than PFEC~\cite{li2016pruning} (41.1\% v.s. 24.2\%) with comparable accuracy drop.
For pre-trained ResNet-50, SFP leads to 8.27\% top-5 accuracy drop for 41.8\% speedup, but our ASFP could achieve negligible top-5 accuracy drop (0.14\%) with the same speedup ratio.
The maintained model capacity and asymptotic pruning of ASFP are the main reasons for the improved accuracy and efficiency.

%%%%%%%%%%%%%%%%%%%%%%%%%%%%%%%%%%%%%%%%%%%%%%%%%%%%%%%%%%%%%%%%%%%%%%%%%%%%%%%%%%%%%%%%%%%%%%%%%%%

%%%%%%%%%%%%%%%%%%%%%%%%%%%%%%%%%%%%%%%%%%%%%%%%%%%%%%%%%%%%%%%%%%%%%%%%%%%%%%%%%%%%%%%%%%%%%%%%%%%

\textbf{Realistic Acceleration.} 
In order to test the realistic speedup ratio, we measure the forward time of the pruned models on one GTX 1080 Ti GPU with a batch size of 64. The results are shown in Table~\ref{table:Comparison_Speed}.
\BB{The gap between theoretical and realistic speed may come from non-tensor layers and the limitation of IO delay, buffer switch and the efficiency of BLAS libraries~\cite{Dong_2017_CVPR}.}

\subsection{Comparing SFP and ASFP}

\begin{figure*}[ht]
\begin{minipage}{.64\textwidth}
\centering
\subfigure[SFP on ResNet-18]{
\includegraphics[width=0.45\linewidth]{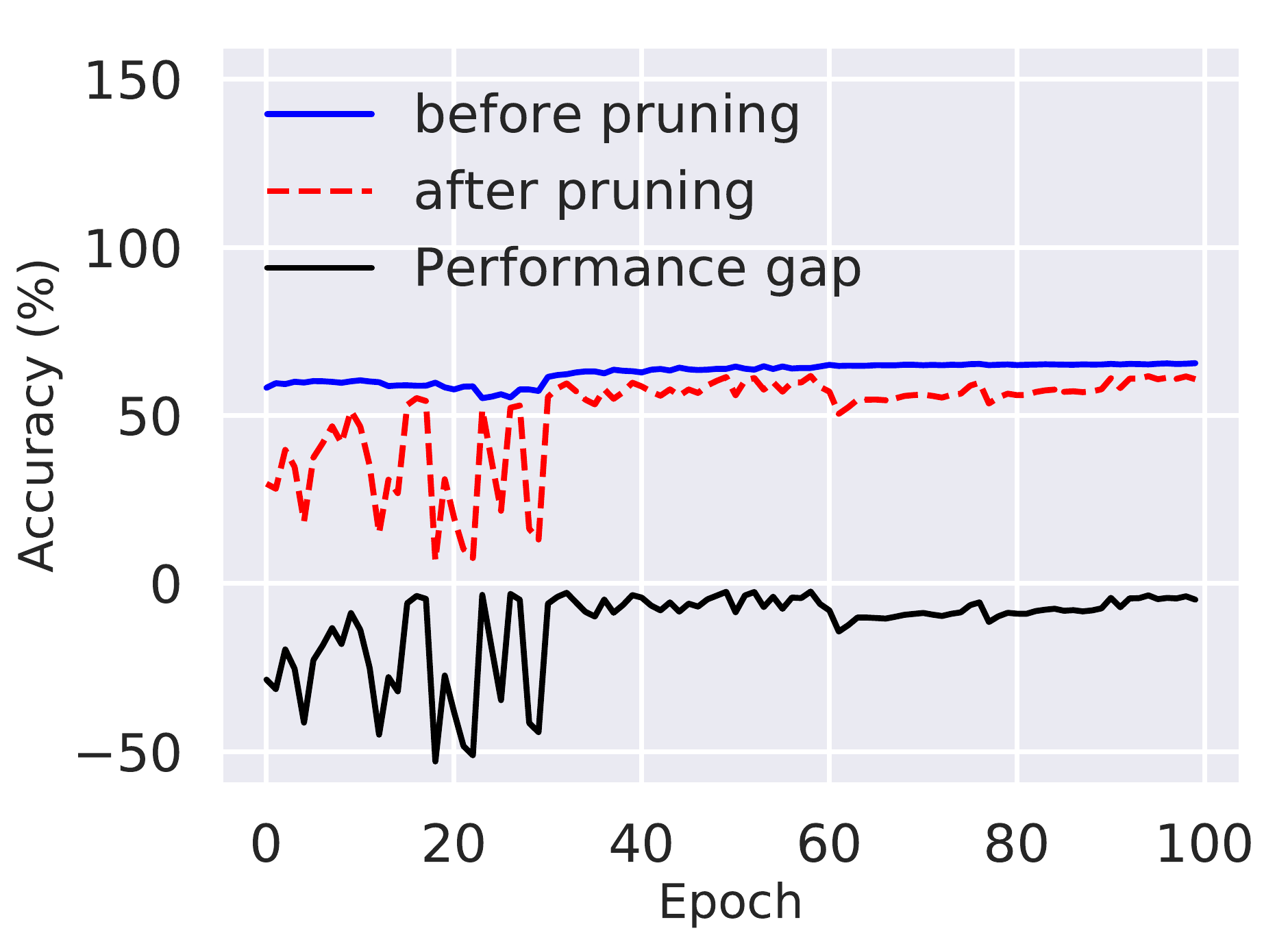}\label{fig:SFP_res18}
}
\subfigure[ASFP on ResNet-18]{
\includegraphics[width=0.45\linewidth]{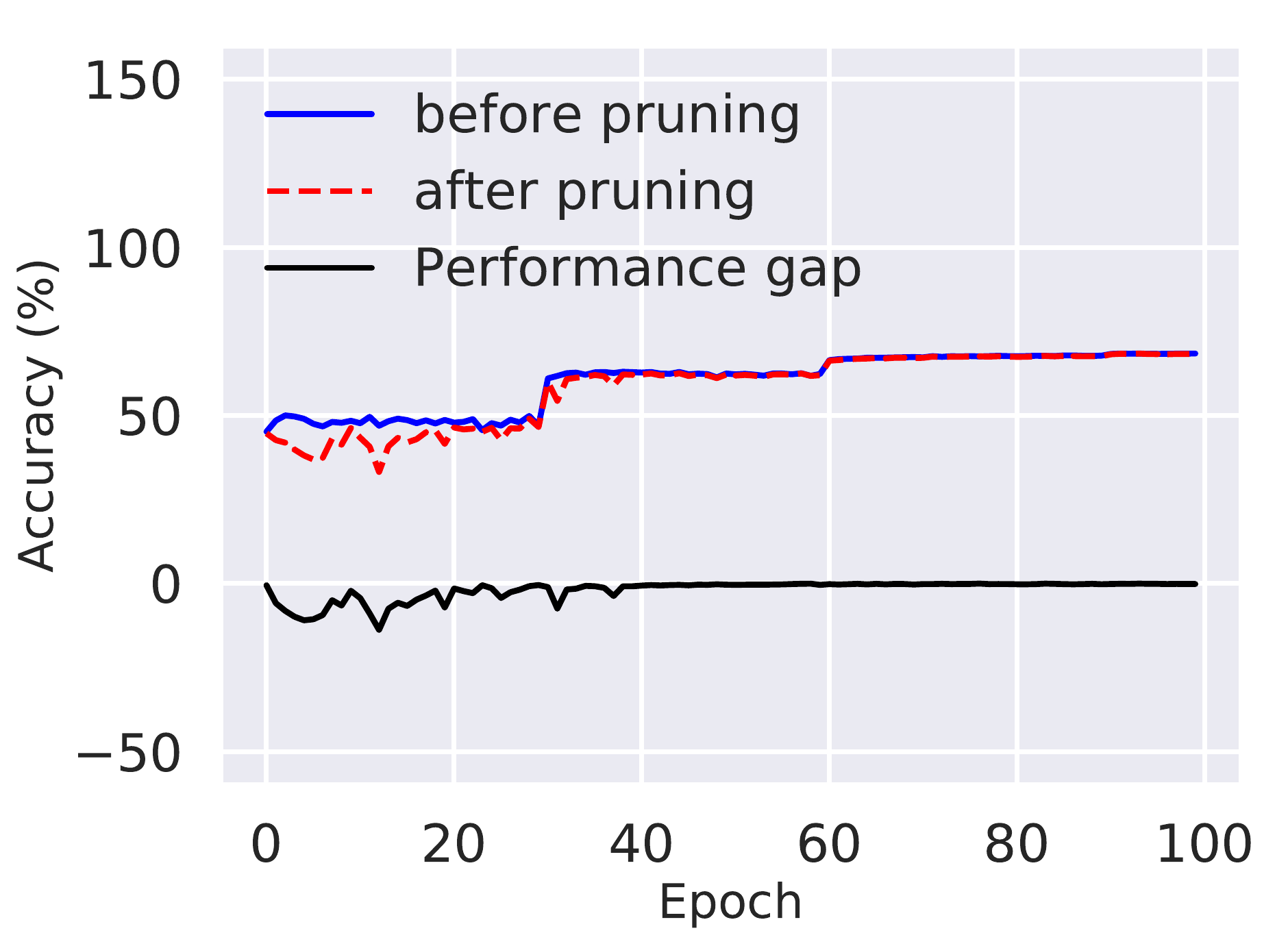}\label{fig:FPF_res18}
}
\subfigure[SFP on ResNet-50]{
\includegraphics[width=0.45\linewidth]{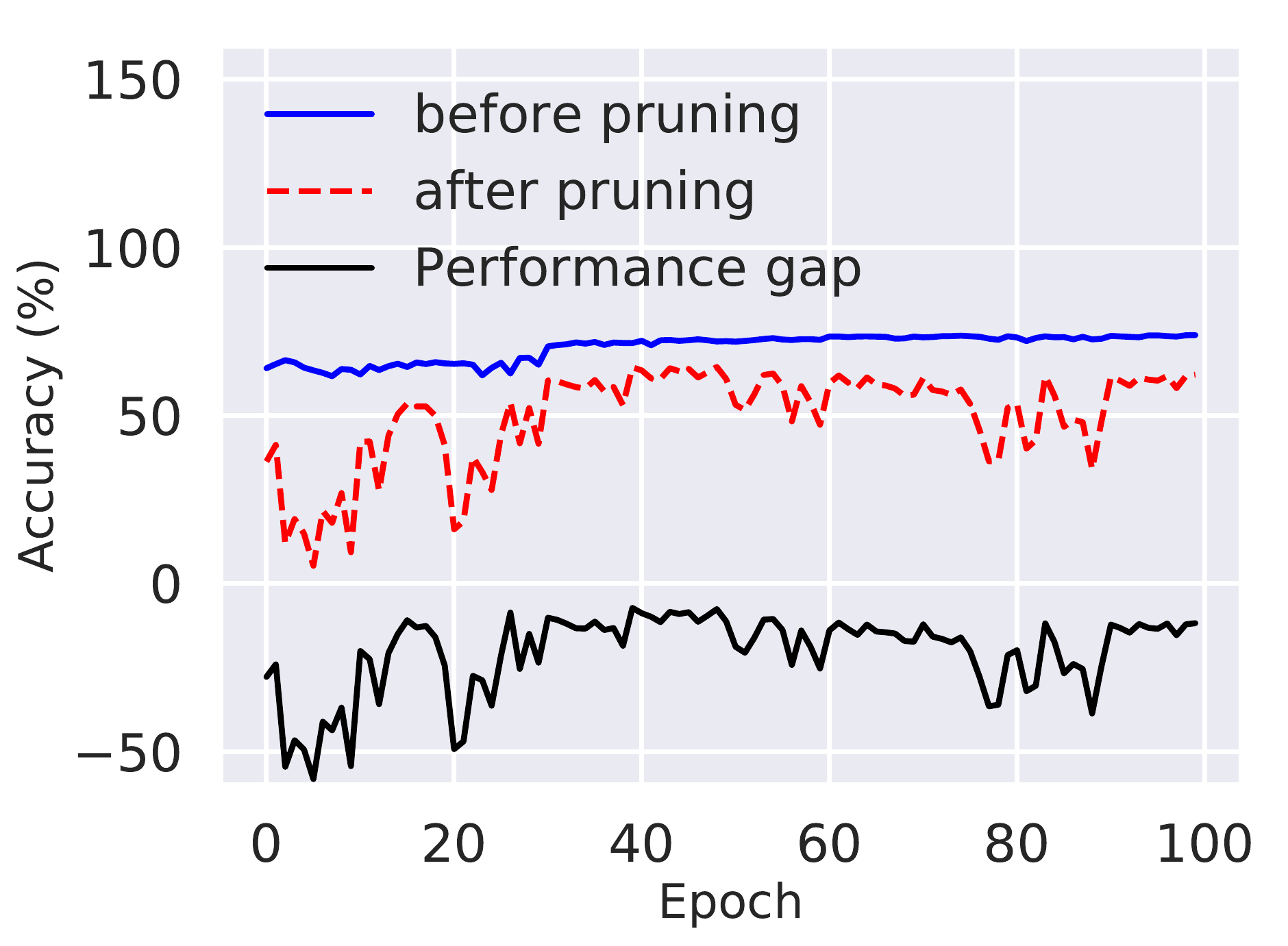}\label{fig:SFP_res50}
}
\subfigure[ASFP on ResNet-50]{
\label{fig:PFP_res50}
\includegraphics[width=0.45\linewidth]{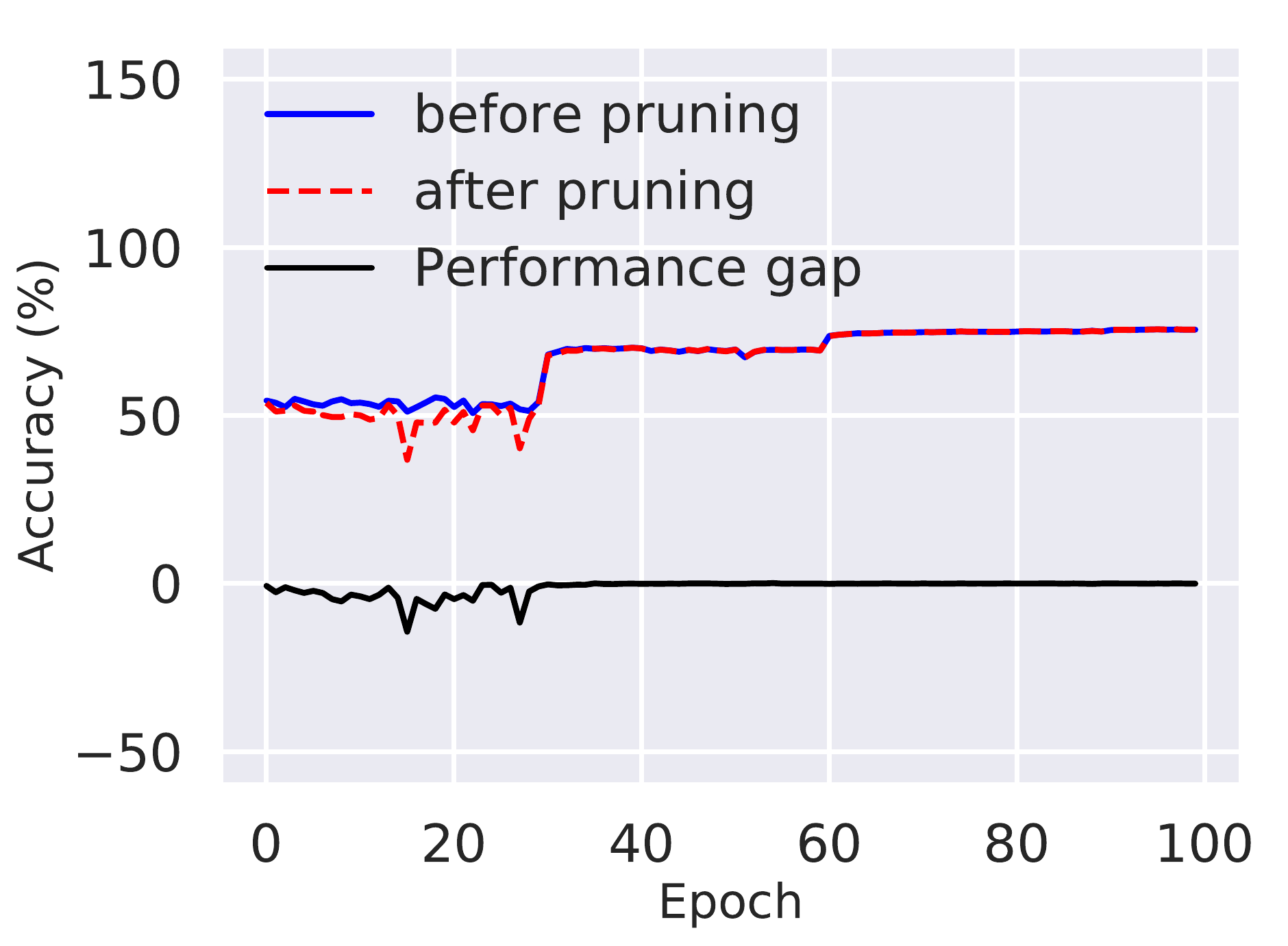}
}
\caption{
The training process of ResNet-18 and ResNet-50 and on ImageNet regarding SFP and ASFP. The solid blue line and red dashed line indicate the accuracy of the model before and after pruning, respectively. The black line is the performance gap due to pruning, which is calculated by the accuracy after pruning subtracting that before pruning. 
}\label{fig:performance_gap}
\end{minipage}
\hfill
\begin{minipage}{0.33\textwidth}
\subfigure[Different pruning intervals.]{
\label{fig:different_epoch}
\includegraphics[width=0.95\linewidth]{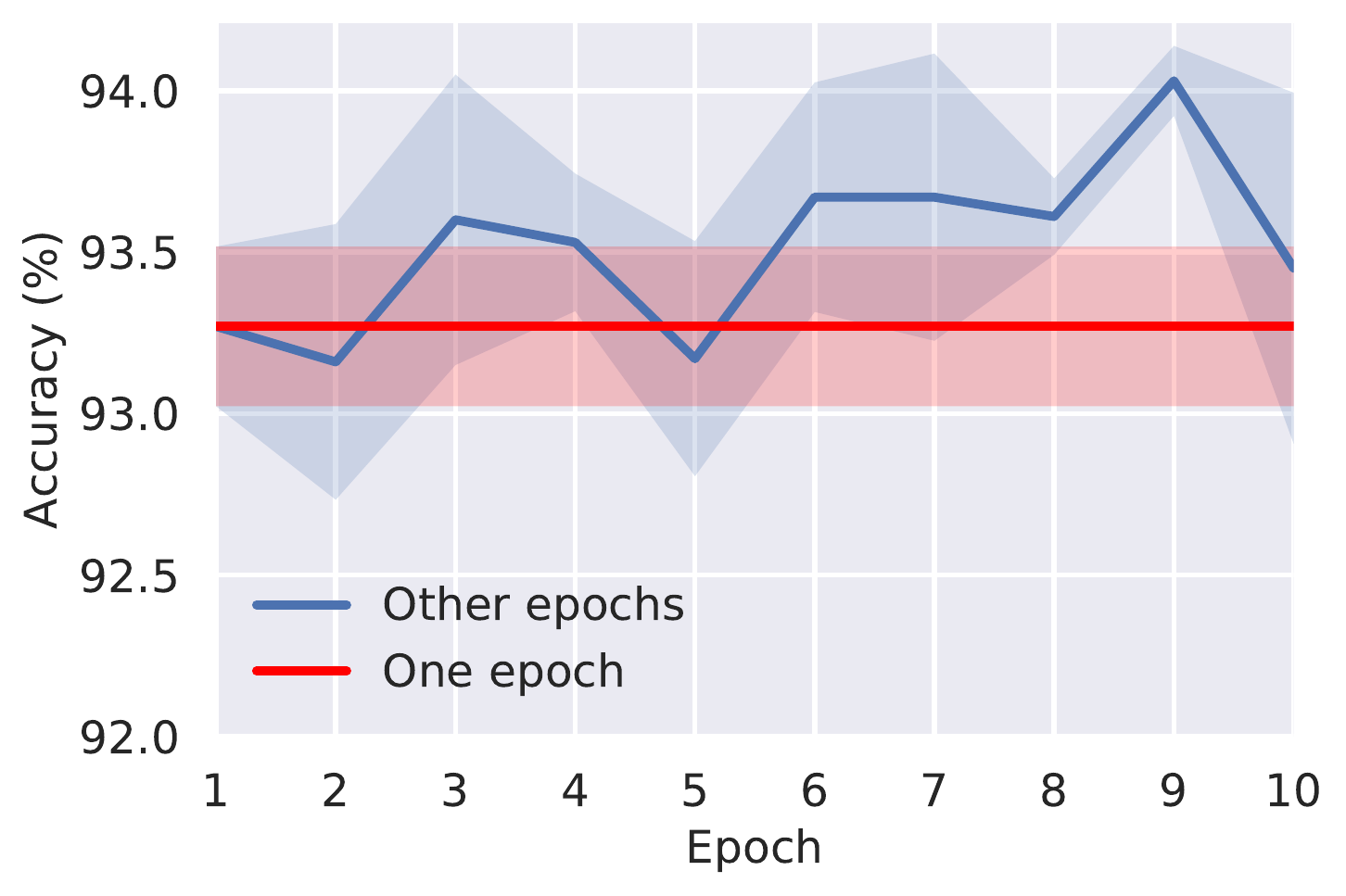}
}
\subfigure[Different parameter D in the Eq.~\ref{eq:progress_rate}.]{
\label{fig:parameter_D}
\includegraphics[width=0.99\linewidth]{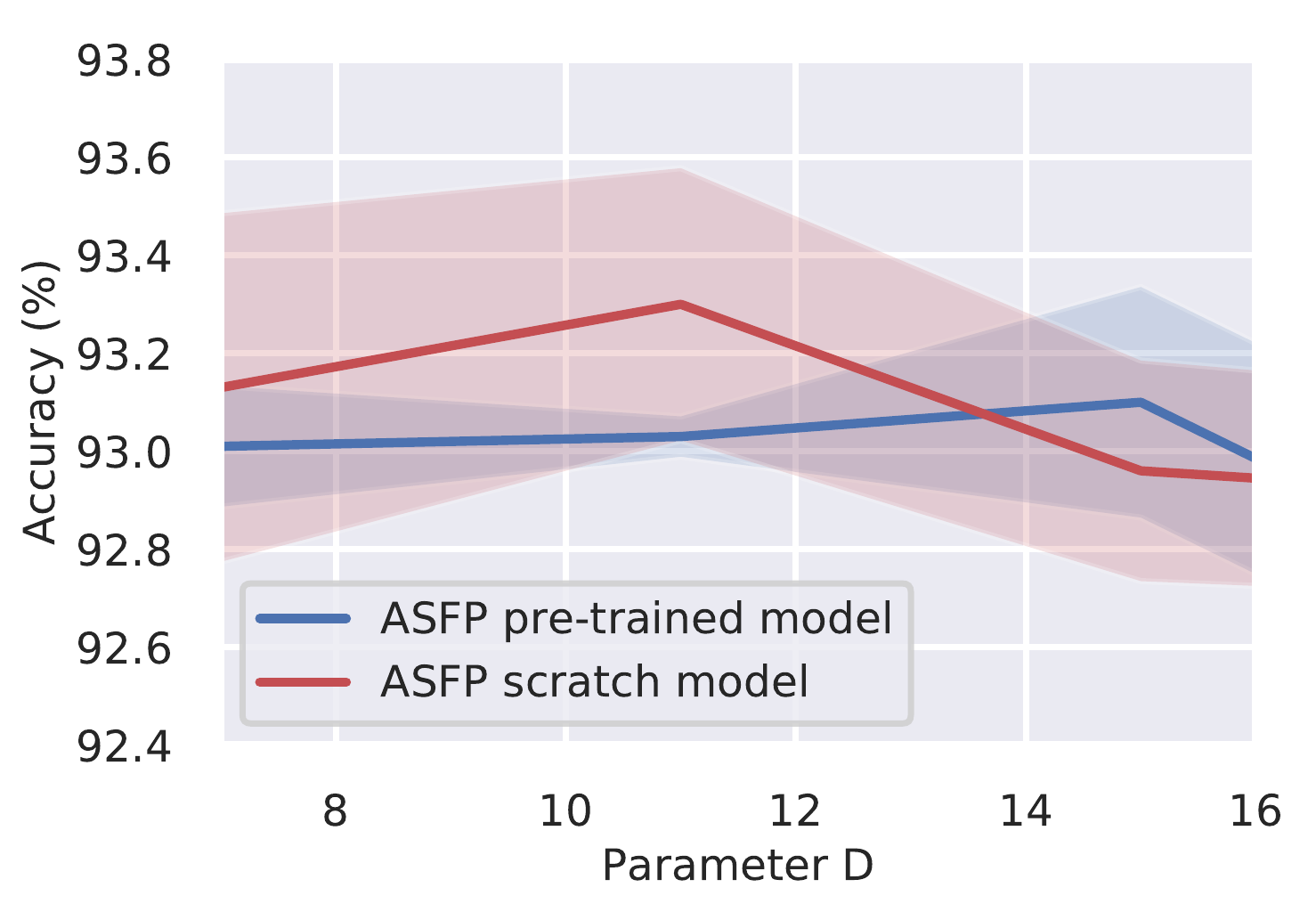}
}
\caption{Ablation study of ASFP. 
(Solid line and shadow denote the mean and standard deviation of three experiments, respectively.)
}\label{fig:ablation}
\end{minipage}
%\vspace{-2mm}
\end{figure*}
%\colorbox{blue!30}{\parbox{1\linewidth}{

\textbf{Performance Regrading Ratio of Pruned FLOPs.}
In Figure~\ref{fig:different_rate}, we test the accuracy of ResNet-56 and ResNet-110 under different ratios of pruned FLOPs.
For pruning pre-trained initialization, as shown in Figure~\ref{fig:different_rate_a} and Figure~\ref{fig:different_rate_b}, ASFP could obtain better performance than SFP on almost all ratio of pruned FLOPs.
Even for pruning models with the random initialization, as shown in Figure~\ref{fig:different_rate_c}, our method could still outperform SFP.
All the results verify that ASFP provides a more effective way to reduce the information loss and thus improves the network performance.

\textbf{Stable Training Process of ASFP.} 
The model accuracies during training for SFP and ASFP are shown in Figure~\ref{fig:performance_gap}. We run this comparison experiment on ResNet-18 and ResNet-50, and the pruning rate is 30\%.
We find that the performance gap is not stable for SFP during almost all the 100 retraining epochs. 
On the contrary, the performance gap of ASFP is much more stable than that of SFP. 
For SFP, directly pruning a large number of filters leads to severe information loss. It is difficult for the network to recover from this, and it leads to an unstable training process.
In contrast, ASFP would result in a small amount of information loss, consequently increasing the stability of the pruning process.

%%%%%%%%%%%%%%%%%%%%%%%%%%%%%%%%%%%%%%%%%%%%%%%%%%%%%%%%%%%%%%%%%%%%%%%%%%%%%%%%%%%%%%%%%%%%%%%%%%%

\subsection{Ablation Study}
Extensive ablation study is also conducted to further analyze each
component of our model.\label{1-norm and 2-norm}

 \subsubsection{Filter Selection Criteria}
The magnitude based criteria such as $\ell_{p}$-norm are widely used to filter selection because computational resources cost is small~\cite{li2016pruning}.
We compare the $\ell_{2}$-norm and $\ell_{1}$-norm, and the results are shown in Table~\ref{table:2norm}.
We find that the performance of $\ell_{2}$-norm criteria are slightly better than that of $\ell_{1}$-norm criteria.
The result of $\ell_{2}$-norm is dominated by the largest element, while the result of $\ell_{1}$-norm is also largely affected by other small elements.
Therefore, filters with some large weights would be preserved by the $\ell_{2}$-norm criteria.
Consequently, the corresponding discriminative features are kept so the accuracy of the pruned model is better.

% \textbf{Filter Selection Criteria.}
% %\label{1-norm and 2-norm}
% Filter selection criteria such as magnitude~\cite{li2016pruning},
% entropy~\cite{luo2017entropy} and Taylor expansion~\cite{molchanov2016pruning}
% have been reported. However, the calculation of entropy and Taylor
% expansion method will cost much more computational resources than
% the magnitude based criteria, so we choose magnitude based criteria.

% We compare the $\ell_{2}$-norm and $\ell_{1}$-norm selection criteria
% and the result is shown in Table~\ref{table:2norm}. We find that
% the performance of $\ell_{2}$-norm criteria is slightly better than that of
% $\ell_{1}$-norm criteria. The result of $\ell_{2}$-norm is dominated by the
% largest element, while the result of $\ell_{1}$-norm is also largely
% affected by other small elements. Therefore, a filter with some large
% weights would be preserved in the $\ell_{2}$-norm criteria. Consequently,
% the corresponding discriminative features are kept so the performance
% of the pruned model is better. 

\setlength{\tabcolsep}{0.5em}  
%{\renewcommand{\arraystretch}{1.1}
\begin{table}[!ht] 
\centering 	
\caption{
Accuracy of CIFAR-10 on ResNet-110 under different pruning rate with different filter selection criteria.
}
\begin{tabular}{c|c c c c } 		  \hline
Pruning rate(\%)   &   10             & 20   &30   \\   \hline        
$\ell_1$-norm      & 93.68 $\pm$ 0.60 &93.68 $\pm$ 0.76  &93.34 $\pm$ 0.12 \\ 		
$\ell_2$-norm      & \textbf{93.89 $\pm$ 0.19} &\textbf{93.93 $\pm$ 0.41} &\textbf{93.38 $\pm$ 0.30}\\ 		     \hline
\end{tabular}  	

\label{table:2norm}
\end{table}

%%%%%%%%%%%%%%%%%%%%%%%%%%%%%%%%%%%%%%%%%%%%%%%%%%%%%%%%%%%%%%%%%%%%%%%%%%%%%%%%%%%%%%%%%%%%%%%%%%%

%%%%%%%%%%%%%%%%%%%%%%%%%%%%%%%%%%%%%%%%%%%%%%%%%%%%%%%%%%%%%%%%%%%%%%%%%%%%%%%%%%%%%%%%%%%%%%%%%%%

\subsubsection{Varying Pruned FLOPs}
We evaluate the accuracy of different pruned FLOPs for ResNet-110, and show the results in Figure~\ref{fig:different_rate}.
When the ratio of pruned FLOPs is less than 40\%, ASFP and SFP achieve similar accuracy. However, when the ratio of pruned FLOPs is more than 40\%, ASFP could obtain much better performance than SFP. 
This is because pruning a large number of filters leads to severe information lose and ASFP is especially effective for this case. In contrast, when only a small portion of the information is lost, the maintained model capacity of SFP is enough for good results.
For the pruning rate between 0\% and about 23\%, the accuracy
of the accelerated model is higher than the baseline model.
This shows that our ASFP and SFP both have a regularization effect on the neural network.

\subsubsection{Selection of the Pruned Layers}
Previous work always prunes a portion of the layers of the network. 
Besides, different layers always have different pruning rates. For example,~\cite{li2016pruning} only prunes insensitive layers, ~\cite{Luo_2017_ICCV} skips the last layer of every block of the ResNet, and~\cite{Luo_2017_ICCV} prunes more aggressively for shallower layers and prune less for deep layers.

Similarly, we compare the performance of pruning the first and second layer of all basic blocks of ResNet-110. We set the pruning rate as 30\%. The model with all the first layers of blocks pruned has an accuracy of $93.96\pm0.13\%$, while the model with all the second layers of blocks pruned has an accuracy of $93.38\pm0.44\%$.
If we carefully select the pruned layers based on the sensitivity, performance improvement may be potentially obtained. However, tuning these hyper-parameters is not the focus of this manuscript.

\subsubsection{Sensitivity of the ASFP Interval}
By default, we conduct our ASFP operation at the end of every training
epoch, we call the ASFP interval equals one under this setting.
However, different ASFP intervals may lead to a different performance, so we explore the sensitivity of ASFP interval. 
We use ResNet-110 under a pruning rate of 30\% as a baseline, and change the ASFP interval from one epoch to ten epochs. The result is shown in Figure~\ref{fig:different_epoch}. 
We find the model accuracy of most (80\%) intervals surpasses the accuracy of one epoch interval.
Therefore, we can even achieve better performance if we fine-tune this parameter.

 \subsubsection{Sensitivity of Parameter D of ASFP}

We change the parameter D in the Eq.~\ref{eq:progress_rate} to comprehensively understand ASFP, and the results are shown in the Figure~\ref{fig:parameter_D}. 
We prune scratch and pre-trained ResNet-56 on CIFAR-10 and set the pruning rate as 40\%. When changing the parameter D from 7 to 16, we find the model accuracy has no large fluctuation ($<0.3\%$).
This shows that the final result of pruning is not sensitive to the parameter D.

%%%%%%%%%%%%%%%%%%%%%%%%%%%%%%%%%%%%%%%%%%%%%%%%%%%%%%%%%%%%%%%%%%%%%%%%%%%%%%%%%%%%%%%%%%%%%%%%%%%

\section{Conclusion \& Future Work}

In this paper, we propose an asymptotic soft filter pruning approach (ASFP) to accelerate the deep CNNs.
As the training procedures go, we allow the pruned filters to be updated and asymptotically adjust the pruning rate.
The soft manner could maintain the model capacity, and the asymptotic pruning could make the pruning process more stable.
Therefore, our ASFP could achieve superior performance. 
Remarkably, without using the pre-trained model, our ASFP can achieve competitive performance compared to the state-of-the-art approaches. Moreover, by leveraging the pre-trained model, our ASFP achieves better results.

%Although changing the pruning rate asymptotically would be beneficial, it is worthwhile to investigate a better way to change the pruning rate. Besides, it is interesting to explore a new way to evaluate the importance of filters.

Although changing the pruning rate asymptotically would be beneficial, it has several limitations. First, several additional hyper-parameters are necessary to define the pruning rate during training. Second, the schedule of the pruning rate is hand-crafted and may not be the best schedule. Third, the theoretical demonstration of the information loss of the network after pruning need to be investigated. 
Furthermore, ASFP can be combined with other acceleration algorithms, \textit{e.g.}, matrix decomposition and low-precision weights, to further improve the performance. 
We will explore these directions in our future work.

%%%%%%%%%%%%%%%%%%%%%%%%%%%%%%%%%%%%%%%%%%%%%%%%%%%%%%%%%%%%%%%%%%%%%%%%%%%%%%%%%%%%%%%%%%%%%%%%%%%

%\section*{Acknowledgment}


\ifCLASSOPTIONcaptionsoff
  \newpage
\fi



%\bibliographystyle{IEEEtran}
\bibliographystyle{IEEEtran}
% \bibliography{IEEEfull,ref}
\bibliography{IEEEabrv,ref}

%\begin{thebibliography}{1}

%H.~Kopka and P.~W. Daly, \emph{A Guide to \LaTeX}, 3rd~ed.\hskip 1em plus 0.5em minus 0.4em\relax Harlow, England: Addison-Wesley, 1999.

%\end{thebibliography}


\begin{IEEEbiography}
[{\includegraphics[width=1in,height=1.25in,clip,keepaspectratio]{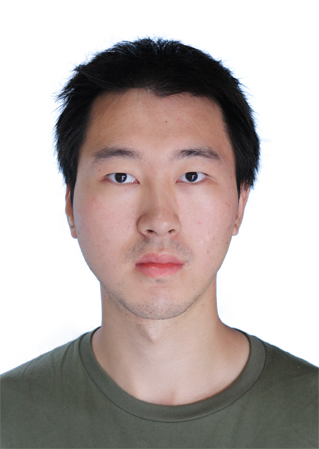}}]
{Yang He}
received the BS degree and MSc from University of Science and Technology of China, Hefei, China, in 2014 and 2017, respectively. He is currently working toward the PhD degree with Center of AI, Faculty of Engineering and Information Technology, University of Technology Sydney. His research interests include deep learning, computer vision, and machine learning.
\end{IEEEbiography}

\begin{IEEEbiography}
[{\includegraphics[width=1in,height=1.25in,clip,keepaspectratio]{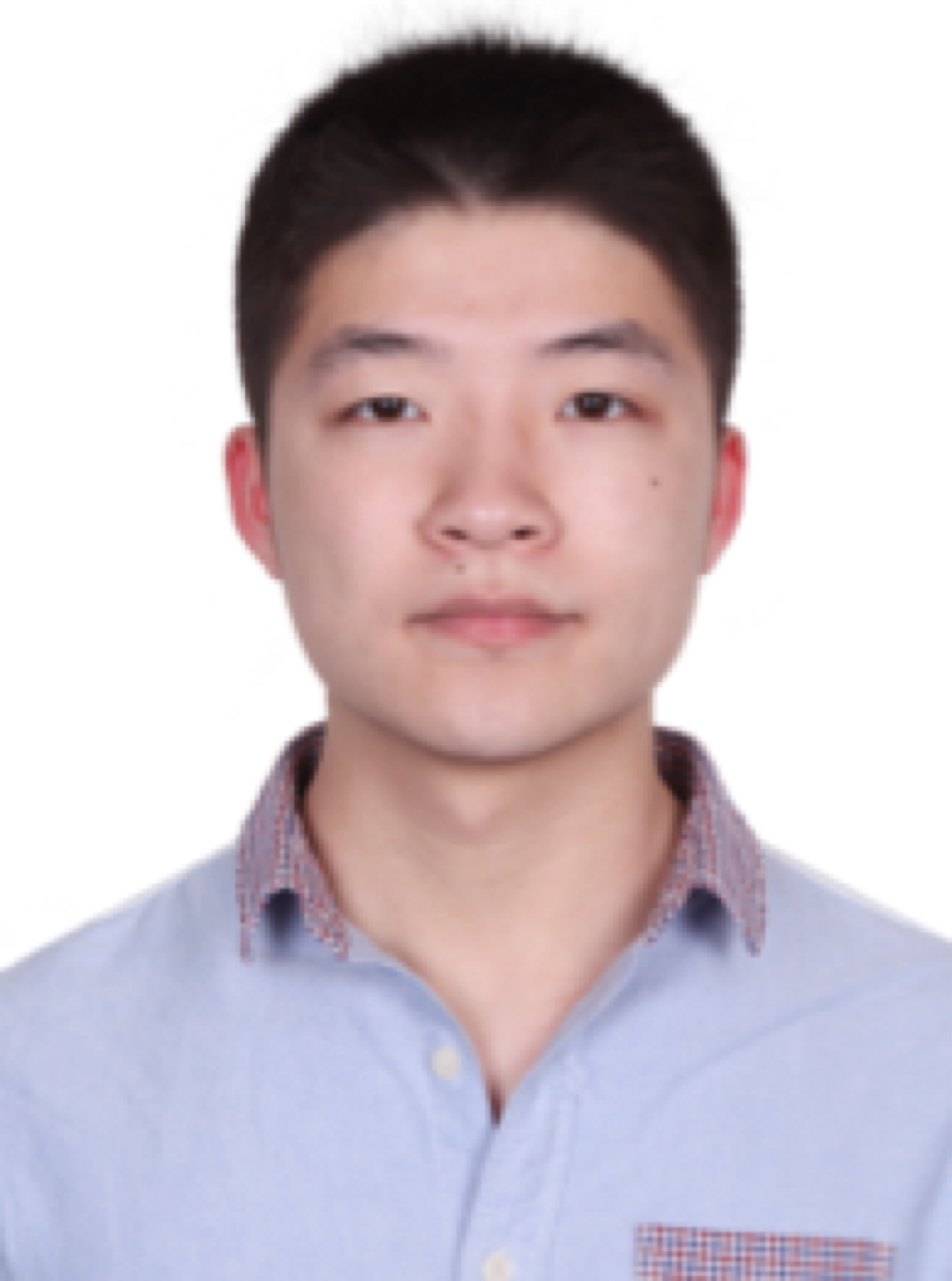}}]
{Xuanyi Dong} received the B.E. degree in Computer Science and Technology from Beihang University, Beijing, China, in 2016.
He is currently a Ph.D. student in the Center for Artificial Intelligence, University of Technology Sydney, Australia, under the supervision of Prof. Yi Yang.
\end{IEEEbiography}

\begin{IEEEbiography}
[{\includegraphics[width=1in,height=1.25in,clip,keepaspectratio]{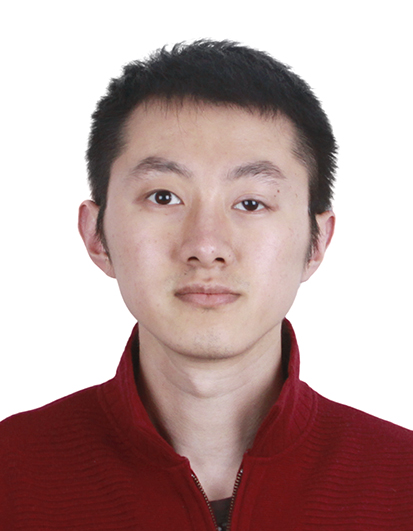}}]
{Guoliang Kang}
received the BS degree in automation from Chongqing University, Chongqing, China, in 2011 and the MSc degree in pattern recognition and intelligent system from Beihang University, Beijing, China, in 2014. He is currently working toward the PhD degree with Center of AI, Faculty of Engineering and Information Technology, University of Technology Sydney. His research interests include deep learning, computer vision, and statistical machine learning.
\end{IEEEbiography}

\begin{IEEEbiography}[{\includegraphics[width=1in,height=1.25in,clip,keepaspectratio]{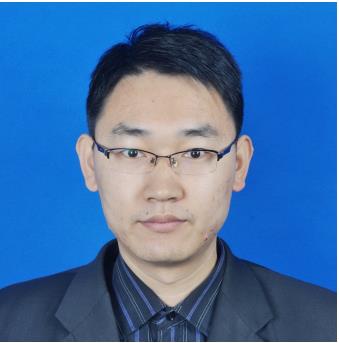}}]{Yanwei Fu}
received the PhD degree from Queen Mary University of London in 2014, and
the MEng degree in the Department of Computer Science \& Technology at Nanjing University in 2011, China. He worked as a Post-doc in Disney Research at Pittsburgh from 2015-2016. He is currently an Assistant Professor at Fudan University. His research interest is image and
video understanding, and life-long learning.
\end{IEEEbiography}

\begin{IEEEbiography}
[{\includegraphics[width=1in,height=1.25in,clip,keepaspectratio]{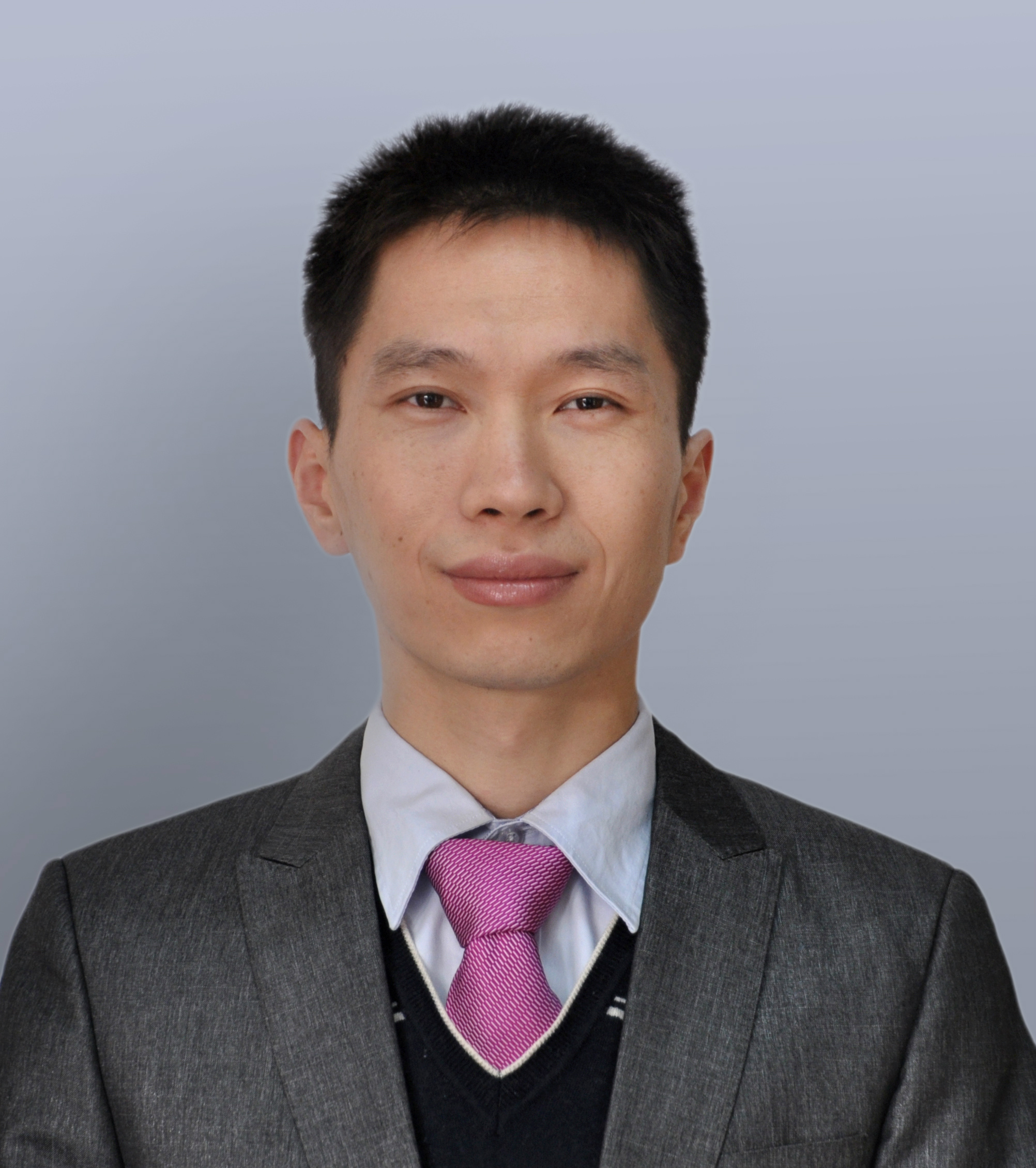}}]
{Chenggang Yan}
received the B.S. degree from Shandong University, Jinan, China, in 2008, and the Ph.D. degree from the Institute of Computing Technology, Chinese Academy of Sciences, Beijing,China, in 2013, both in computer science.
\newline
He is currently a Professor with Hangzhou Dianzi University, Hangzhou, China. Before that, he was an Assistant Research Fellow with Tsinghua University. His research interests include machine learning, image processing, computational biology, and computational photography. He has authored or co-authored more than 30 refereed journal and conference papers. As a co-author, Prof. Yan was the recipient of the Best Paper Awards in the International Conference on Game Theory for Networks 2014, and SPIE/COS Photonics Asia Conference9273 2014, and the Best Paper Candidate in the International Conference on Multimedia and Expo 2011.
\end{IEEEbiography}

\begin{IEEEbiography}
[{\includegraphics[width=1in,height=1.25in,clip,keepaspectratio]{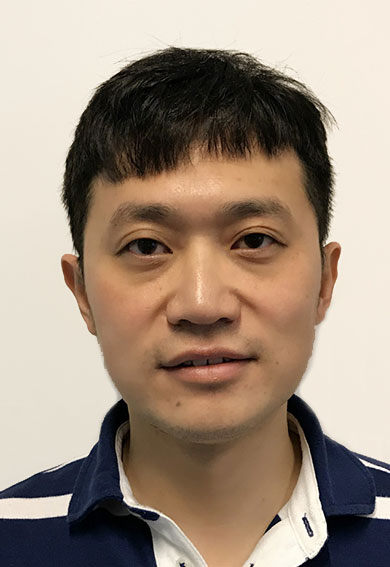}}]
{Yi Yang}
received the Ph.D. degree in computer science from Zhejiang University, Hangzhou, China, in 2010. He is currently a professor with University of Technology Sydney, Australia. He was a Post-Doctoral Research with the School of Computer Science, Carnegie Mellon University, Pittsburgh, PA, USA. His current research interest include machine learning and its applications to multimedia content analysis and computer vision, such as multimedia indexing and retrieval, surveillance video analysis and video semantics understanding.
\end{IEEEbiography}

%\newpage


%\vfill


\end{document}